\documentclass[12pt]{iopart}

\usepackage{cite}

\usepackage{graphicx}
\usepackage{subcaption}
\usepackage{amsmath}
\usepackage{geometry}
\usepackage{hyperref}

\usepackage{algorithm}
\usepackage{algorithmic}

\usepackage{siunitx}

\begin{document}

\title{Evaluation of uncertainty estimations for Gaussian process regression based machine learning interatomic potentials}

\author{Matthias Holzenkamp$^*$, Dongyu Lyu$^\dagger$, Ulrich Kleinekath\"ofer$^\dagger$ and Peter Zaspel$^*$}

\address{$^*$ School of Mathematics and Natural Sciences, Bergische Universit\"at Wuppertal, Gau{\ss}stra{\ss}e 20, 42119 Wuppertal, Germany}
\address{$^\dagger$ School of Science, Constructor University, Campus Ring 1, 28759 Bremen, Germany}

\ead{zaspel@uni-wuppertal.de}
\date{
}

\begin{abstract}
Uncertainty estimations for machine learning interatomic potentials (MLIPs) are crucial for quantifying model error and identifying informative training samples in active learning strategies. In this study, we evaluate uncertainty estimations of Gaussian process regression (GPR)-based MLIPs, including the predictive GPR standard deviation and ensemble-based uncertainties. We do this in terms of calibration and in terms of impact on model performance in an active learning scheme. We consider GPR models with Coulomb and Smooth Overlap of Atomic Positions (SOAP) representations as inputs to predict potential energy surfaces and excitation energies of molecules. Regarding calibration, we find that ensemble-based uncertainty estimations show already poor global calibration (e.g., averaged over the whole test set). In contrast, the GPR standard deviation shows good global calibration, but when grouping predictions by their uncertainty, we observe a systematical bias for predictions with high uncertainty. Although an increasing uncertainty correlates with an increasing bias, the bias is not captured quantitatively by the uncertainty. Therefore, the GPR standard deviation can be useful to identify predictions with a high bias and error but, without further knowledge, should not be interpreted as a quantitative measure for a potential error range. Selecting the samples with the highest GPR standard deviation from a fixed configuration space leads to a model that overemphasizes the borders of the configuration space represented in the fixed dataset. This may result in worse performance in more densely sampled areas but better generalization for extrapolation tasks.
\end{abstract}

\section{Introduction}
Machine learning (ML) models that map atomic structures to corresponding properties have seen a significant increase in research interest over the past decade. These models promise to combine the accuracy of ab initio calculations with the speed, and scalability to large systems, of empirical potentials. Of particular interest are models that predict potential energies and possibly the associated forces, known as machine learning interatomic potentials (MLIPs). The idea is to perform highly accurate reference calculations, e.g.~with density functional theory, for only a few molecular configurations. These samples then serve as training data for a machine learning model, which can predict new samples at substantially lower costs.

How well samples from different areas of the configuration space are predicted by an MLIP depends highly on its training data. Samples from areas of the configuration space (e.g., spatial arrangements of atoms in an atomic system) that differ too much from those covered by the training data might not be predicted well. Because of this variability in prediction accuracy, it becomes desirable to provide an uncertainty estimation for every model prediction. A good uncertainty estimation could help to understand the range of a potential error of a prediction, and be utilized in active learning (AL) strategies\cite{rupp2014machine, chen2023accelerating, flare, van2023hyperactive, montes2022training, kulichenko2023uncertainty, podryabinkin2017active, zhang2019active, uteva2018active, zhu2023fast, zaverkin2022exploring, wilson2022batch, toyoura2016machine, sivaraman2020machine, lin2020automatically, guan2018construction, schwalbe2021differentiable}. With AL, algorithms are referred to that aim to choose or create new training samples, which are maximally informative to the model.

Many different ML approaches to design MLIPs exist. One class of MLIPs are message passing neural networks \cite{schnet, painn, nequip, mace} (MPNN), which contain features that are trained via message passing steps. In other approaches, instead of trained features, fixed representations are employed that either globally represent the whole system of atoms \cite{coulomb, mbtp, bag_of_bonds} or locally represent atomistic environments \cite{soap, snap, ace, behler_parrinello}. These representations provide important properties such as invariance against translation, rotation, and atom permutation\cite{langer2022representations}. Besides linear models \cite{snap, linear_ace} and neural networks\cite{behler_parrinello, gaussian_moments, ani}, kernel-based models are commonly chosen as ML model \cite{gap, coulomb, bag_of_bonds}. Within kernel-based models, many are variations of Gaussian process regression\cite{williams2006gaussian} (GPR). GPR predicts the output for a new input by generating a Gaussian predictive distribution, where the mean represents the (deterministic) prediction, and the standard deviation provides a measure of uncertainty for the prediction. The GPR standard deviation is commonly used as an uncertainty measure in AL schemes for GPR-based MLIPs\cite{uteva2018active, guan2018construction, flare, rupp2014machine}. Besides the GPR standard deviation, other uncertainty estimations are conceivable for GPR. A common approach in ML is to construct ensembles of slightly different models and use the standard deviation of their predictions as an uncertainty measure. Since GPR is a non-parametric method that relies directly on the training data for predictions, it is reasonable to construct ensembles where the models differ based on subsets of the training data. Uteva et al.\cite{uteva2018active} trained two models on half of the initial training set each. Similarly, a bootstrap aggregation\cite{bootstrap, carrete2023deep} ensemble, which is commonly used in ML, can be applied to GPR.

Both the GPR standard deviation and ensemble-based uncertainties are standard deviations of predictive distributions, thus provide uncertainty estimations. To evaluate these uncertainty estimations, the most important aspect is calibration, which describes how well the distribution of empirical errors aligns with the predictive distributions. A common tool to evaluate calibration are calibration curves\cite{kuleshov2018accurate}, which test whether the empirical errors are covered by the confidence intervals derived from the predictive distributions. However, it has been pointed out that this approach does not sufficiently validate good calibration, because it averages over the entire test dataset \cite{pernot2022prediction, levi2022evaluating}. Separating predictions into bins by their uncertainty can give a more localized evaluation of uncertainty estimations, for example, using reliability diagrams\cite{levi2022evaluating}, or by calculating the variance of z-scores\cite{pernot2022prediction}. In the literature evaluating uncertainty estimations of general MLIPs, local analyses of calibration are often lacking or limited to qualitative tests, such as plotting errors against uncertainties \cite{tran2020methods, kellner2024uncertainty, zhu2023fast, peterson2017addressing}. In contrast, Busk et al. \cite{busk2023graph} sorted all predictions into bins to generate reliability diagrams and calculate local z-score variances. In this work, we use calibration curves as well as \textit{extended reliability diagrams} to evaluate calibration. While classical reliability diagrams only show the standard deviation of empirical errors within each bin, extended reliability diagrams additionally show the mean of empirical errors. This accounts for the fact that, when only the standard deviation of errors is shown, a systematical shift of the error distribution is not visible.

Beside explicitly analyzing calibration, uncertainty estimations can also be evaluated by their performance when utilized in AL schemes. While a wide variety of AL schemes exist in the MLIP literature, in this work we restrict ourselves to a simple and widely used AL approach from standard machine learning literature, known as uncertainty sampling \cite{uncertainty_sampling, uteva2018active}. Uncertainty sampling samples a fixed configuration space, represented in a fixed dataset. An initial model is trained on a small random subset of the dataset, and the uncertainty estimation of this model is then used to iteratively select samples from the remaining pool. In the general ML literature, uncertainty sampling is often applied to classification problems, but it is also used for regression problems \cite{mussmann2018relationship}. It was shown that uncertainty sampling can improve data efficiency, which refers to the reduction in the number of training samples needed to achieve comparable model accuracy compared to using randomly selected samples, by a multiple. However, the extent of this improvement strongly depends on the dataset and the uncertainty measure used \cite{mussmann2018relationship}. In this study, uncertainty sampling is applied to analyze the benefit of the discussed different uncertainty measures on AL.

When it comes to GPR-based MLIPs, the calibration of uncertainty estimations has rarely been evaluated explicitly. Aside from qualitative visualizations of calibration, such as plotting errors against uncertainties\cite{flare, bartok2022improved}, Tran et al. \cite{tran2020methods} evaluated different aspects of uncertainty estimations for an exemplary dataset using, among others, a GPR-based MLIP. More often, uncertainty estimations of GPR-based MLIPs are evaluated by directly examining the performance when applied to AL, without explicitly analyzing calibration\cite{flare, uteva2018active, rupp2014machine, guan2018construction}. However, these studies give a mixed picture on the effect of AL on training GPR-based MLIPs, and it is not always clear whether AL strategies provide an actual improvement over random sampling. 

The objective of this study is to provide a comprehensive analysis of uncertainty estimations for GPR-based MLIPs. To achieve this, we provide an explicit evaluation of calibration and investigate the performance of different uncertainty measures applied to active learning (AL). We analyze global calibration using calibration curves and local calibration of groups of predictions, separated by their uncertainty, with extended reliability diagrams. Uncertainty sampling is used as a simple AL algorithm. We consider a GPR model with global Coulomb representations\cite{coulomb} as inputs, as well as an atomistic GPR model with SOAP\cite{soap} representations. Our tests include predicting ground-state energies for molecules from well-established benchmarks. Additionally, we predict excited-state energies for a trajectory of porphyrin in the gas phase, chosen as a large and complex molecule with a potentially complex configuration space.

\section{Methods}
\label{sec:methods}
\subsection{Gaussian process regression}
In GPR\cite{williams2006gaussian}, a prior probability distribution over functions is defined using a Gaussian process.
We want to find a posterior predictive distribution from training inputs $\mathbf{x}_i$ and corresponding output measurements $y_i$, with which we can predict outputs for unknown inputs $\textbf{x}$. Our output measurements will always be noisy. This can be considered by assuming that the actual function values differ from the measurements with
\begin{align}
y_i = f\left(\mathbf{x}_i\right) + \epsilon_i,\qquad \text{with}\, \epsilon_i\sim\mathcal{N}\left(0, \sigma_n^2\right)\,,
\end{align}
where $\epsilon_i$ are independent, identically distributed random variables drawn from a Gaussian distribution with zero mean and variance $\sigma_n^2$.
The GPR framework provides a predictive distribution for the output corresponding to any input $\textbf{x}$. The actually predicted value for the output of an input $\mathbf{x}$ is the mean of this predictive distribution, given as
\begin{align}\label{eq:gprMean}
m\left(\mathbf{x}\right) = \mathbf{K}\left(\mathbf{x}, \mathbf{X}\right) \bigl(\mathbf{K}\left(\mathbf{X}, \mathbf{X}\right) + \sigma_n^2\mathbf{I}\bigr)^{-1}\mathbf{y}\,.
\end{align}
The properties of this function are determined by the covariance function $k\left(\mathbf{x}, \mathbf{x'}\right)$, which measures the similarity between two inputs $\mathbf{x}$ and $\mathbf{x'}$. $\mathbf{X}$ is a matrix with the training inputs as rows, $\mathbf{y}$ is a vector with the corresponding outputs. The vector $\mathbf{K}\left(\mathbf{x}, \mathbf{X}\right)$ contains all pairs of kernel evaluations between the input $\mathbf{x}$ and the training inputs $\mathbf{X}$, whereas  $\mathbf{K}\left(\mathbf{X}, \mathbf{X}\right)$ is the matrix of all pairwise kernel evaluations on the training inputs $\mathbf{X}$.\\
In our case, an input $\textbf{x}$ to the model can be either a global representation of the atomistic system or local representations of the atomistic environments. The corresponding output $y$ is a scalar energy value.
For our tests, we chose a Gaussian covariance function
\begin{align}
    k\left(\mathbf{x}, \mathbf{x'}\right) = \sigma_f^2\text{exp}\left(-\frac{1}{2}\left(\mathbf{x} - \mathbf{x}'\right)^T\mathbf{M}\left(\mathbf{x} - \mathbf{x}'\right)\right)\,,
    \label{eq:gaussian_cov}
\end{align}
which is one of the most common choices. We used a different length-scale for every feature, known as automatic relevance determination\cite{williams2006gaussian}, to increase the accuracy and flexibility of the model. Here $\mathbf{M}$ is a diagonal matrix $\text{diag}\left(\frac{1}{l_1^2},...,\frac{1}{l_D^2}\right)$, with $D$ the number of features, and $\sigma_f^2$ is the output-scale. We optimized the hyperparameters  $\sigma_f^2$ and $\sigma_n^2$, as well as length-scales inside the covariance function via maximum marginal log-likelihood, details can be found in the section hyperparameter optimization of the supplementary material. For all our tests, we used the GPR implementation of GPyTorch \cite{gpytorch}.\\
In the case of a global representation as inputs $\mathbf{x}$, the covariance function of Eq.~\ref{eq:gaussian_cov} is directly applied to these inputs. For our study, we incorporated \textit{Coulomb matrices}\cite{coulomb} as a global representation. For Coulomb matrices, to ensure invariance against atom indexing, often a sorting is carried out, which results in discontinuities, which may cause challenges in the construction of ML models \cite{langer2022representations}. In our tests, we ignored invariance against atom permutation and omitted sorting because, for all tasks, a system with fixed atoms in a fixed order was given.\\
As a local representation, we used \textit{Smooth overlap of atomic positions (SOAP)} \cite{soap}. It is commonly used and GPR models with SOAP representations as inputs have shown high accuracy \cite{langer2022representations}. For atomistic representations, an input $\textbf{x}$ will consist of representations $\textbf{x}_j$ for every atomistic environment of the system. The covariance function between two inputs $\textbf{x}$ and $\textbf{x'}$ is built by summing over an atomistic covariance function
\begin{align}
    k\left(\mathbf{x}, \mathbf{x'}\right) = \sum_{j, l=1}^{n_\text{atoms}} k_\text{atomistic}\left(\textbf{x}_j, \textbf{x}'_l\right).
    \label{eq:atomistic_kernel}
\end{align}
between every atomistic representation of $\textbf{x}$ and $\textbf{x'}$ \cite{gap, langer2022representations}. 
The atomistic covariance function\linebreak $k_\text{atomistic}\left(\textbf{x}_j, \textbf{x}'_l\right)$ is the Gaussian covariance function from Eq.~\ref{eq:gaussian_cov}. Because the evaluation of Eq.~\ref{eq:atomistic_kernel} can cause large computational overhead, often a sparsification that only considers a sparse subset of atomistic environments \cite{gap} is applied. However, this approach will also reduce the accuracy. For our tests, we did not use a sparsification. To compute the SOAP representations we utilized the DScribe library\cite{dscribe}. For all tests, we chose a cutoff radius of $r_\text{cut} = 5 \si{\angstrom}$, set the number of radial basis functions to $n_\text{max} = 3$ and the maximum degree of spherical harmonics to $l_\text{max} = 1$. More details on these parameters can be found in \cite{soap}.

\subsection{Uncertainty estimations}
As discussed in the introduction, GPR and ensembles of GPR models provide a probabilistic predictive distribution. In general, we will describe this distribution via a pair
\begin{equation}\label{eq:predictiveDistribution}
\mathcal{D}(m(\mathbf{x}), u(\mathbf{x})^2)\,,
\end{equation}
with a given mean $m(\mathbf{x})$, and a standard deviation $u(\mathbf{x})$, which we consider as uncertainty measure in here. Note that we may interpret the pair from eq.~\ref{eq:predictiveDistribution} as the parameters describing a Gaussian distribution. In fact, the predictive distribution of a GPR model is anyway Gaussian. Moreover, the predictive distributions of GPR ensembles are often assumed to be Gaussian too\cite{busk2023graph, heid2023characterizing}, which is, according to the central limit theorem, at least true for a large number of ensemble models\cite{heid2023characterizing}. 

For the case of the standard GPR model, the predictive distribution is, using notation from eq.~\eqref{eq:predictiveDistribution}, given by a pair
\begin{equation}
\mathcal{D}_{G}(m_{G}(\mathbf{x}),u_{G}(\mathbf{x})^2)\,,
\end{equation}
where $m_{G}(\mathbf{x})$ is the GPR mean from eq.~\ref{eq:gprMean} and we have the \textit{GPR standard deviation}
\begin{align}\label{eq:gprVar}
    u_{G}\left(\textbf{x}\right) = \sqrt{k\left(\mathbf{x}, \mathbf{x}\right) - \mathbf{K}\left(\mathbf{x}, \mathbf{X}\right) \bigl(\mathbf{K}\left(\mathbf{X}, \mathbf{X}\right) + \sigma_n^2\mathbf{I}\bigr)^{-1}\mathbf{K}\left(\mathbf{X}, \mathbf{x}\right)}\,.
\end{align}
Note that the GPR variance is only calculated from the kernel function $k(\cdot,\cdot)$, the \textit{inputs} $\mathbf{X}$ of the training set, i.e.~the sample locations in the feature space, and the noise variance $\sigma_n^2$. Hence it does not depend on the training outputs, i.e.~the trained energies. Intuitively, the GPR variance is high, when $\mathbf{x}$ is far from the training inputs, as measured by the GPR model.
In the variance, see eq.~\eqref{eq:gprVar}, GPR already accounts for the noise $\sigma_n^2$ in the observations, therefore, we do not add this noise to our uncertainty estimation, like it was done by Vandermause et al.\cite{flare}.

The \textit{two set approach},
proposed by Uteva et al.\cite{uteva2018active}, was motivated by the principles underlying the jackknife and bootstrap methods\cite{bootstrap}. We use for its predictive distribution the notation
\begin{equation}
\mathcal{D}_{2}(m_{2}(\mathbf{x}),u_{2}(\mathbf{x})^2)\,.
\end{equation}
This distribution is obtained by splitting the initial training set into two sets which are used to train two GPR models with predictions $m^{\left(1\right)}\left(\textbf{x}\right)$ and $m^{\left(2\right)}\left(\textbf{x}\right)$. The mean over both models then becomes the predictive mean
\begin{equation}
    m_{2}(\mathbf{x}) = \frac{1}{2}(m^{\left(1\right)}\left(\mathbf{x}\right)+m^{\left(2\right)}\left(\mathbf{x}\right))\,,
\end{equation}
while the difference in the predictions of the two models
\begin{align}
    u_{2}\left(\textbf{x}\right) = \left|m^{\left(1\right)}\left(\textbf{x}\right) - m^{\left(2\right)}\left(\textbf{x}\right)\right|\,,
\end{align}
is the uncertainty estimation. The square of this particular uncertainty measure can be interpreted as a very rough estimate of the model's variance with respect to changes in the training set, including the energies. In other words, it very roughly approximates the variance in the bias-variance decomposition of ML models, which is common in ML literature\cite{bishop}.

An extension of this is the \textit{bootstrap aggregation} ensemble\cite{bootstrap}, which leads to the predictive distribution
\begin{equation}
    \mathcal{D}_b = (m_b(\mathbf{x}),u_b(\mathbf{x})^2)\,.
\end{equation}
It is commonly used for neural networks. To calculate the predictive distribution, multiple subsets of the training data are generated by randomly sampling indices from the original dataset, allowing for duplicates. Then, the subsets are used to train an ensemble of multiple models $m^{\left(l\right)}\left(\textbf{x}\right)$. The mean is then 
\begin{equation}
    {m}_b\left(\textbf{x}\right) = \frac{1}{N}\sum_{l=1}^N m^{\left(l\right)}\left(\textbf{x}\right)\,,
\end{equation}
and the uncertainty is calculated by the standard deviation of their different predictions $m^{(l)}\left(\textbf{x}\right)$,
\begin{align}
    {u}_b\left(\textbf{x}\right) = \sqrt{\frac{1}{N-1}\sum_{l=1}^{N-1} \left(m^{\left(l\right)}\left(\textbf{x}\right) - {m}_b\left(\textbf{x}\right)\right)^2}\,,
\end{align}
where $N$ is the number of models. Note that the variance of the bootstrap ensemble is an even better approximation of the variance of the bias-variance decomposition.

As a slight change towards the classical definition of the two-sets and bootstrap approach, we do not use the ensemble mean as mean for their predictive distributions, since we found whether we use the mean of the "full" GPR model or the ensemble mean does not make a difference for the essential results on the correlation between error and uncertainty estimation. A comparison can be found in Fig.~S1 and S2 of the supplementary material.

\subsection{Evaluating uncertainty estimations}
The most important aspect to analyze uncertainty estimations, such as GPR standard deviation, is their relation to the actual prediction error. To better understand, what we could potentially expect from uncertainty estimations in an ideal setting, let us assume an approach that constructs an optimal predictive distribution
$$\mathcal{D}_{opt}(m_{opt}(\mathbf{x}),u_{opt}(\mathbf{x})^2)\,,$$
where the true values $y(\mathbf{x})$ corresponding to predictions $m_{opt}(\mathbf{x})$ actually follow the distribution
$$y(\mathbf{x}) \sim \mathcal{N}(m_{opt}(\mathbf{x}),u_{opt}(\mathbf{x})^2)\,.$$
It is then quite immediate that the prediction error $\epsilon(\mathbf{x})$ of the model for all test samples would behave like
$$\epsilon(\mathbf{x}) = y(\mathbf{x}) - m_{opt}(\mathbf{x}) \sim \mathcal{N}(0,u_{opt}(\mathbf{x})^2)\,.$$
In other words, the error of the model would be normally distributed with a mean of $0$ and a standard deviation of $u_{opt}(\mathbf{x})$. 
Consequently, our subsequent evaluation of the uncertainty estimations aims at testing for a given uncertainty measure $u(\mathbf{x})$, how closely the prediction errors $\epsilon$ would fulfill
\begin{equation}
\epsilon \sim \mathcal{N}(0,u(\mathbf{x})^2))\,.
     \label{eq:error_dist}
\end{equation}

The extent to which the empirical distribution of prediction errors aligns with this ideal distribution is referred to as calibration. A common method for evaluating calibration is the use of calibration curves introduced by Kuleshov et al.\cite{kuleshov2018accurate}. For every prediction within a test set, the uncertainty estimation is used to calculate a confidence interval, providing a range expected to contain the true value with a specific probability $\alpha_\text{predicted}$. By iterating over different $\alpha_\text{predicted}$ values from 0 to 1, the proportion of errors $\alpha_\text{observed}$ falling within the corresponding confidence intervals is calculated and plotted against $\alpha_\text{predicted}$. We use the implementation provided in the uncertainty toolbox\cite{uncertainty_toolbox} to generate calibration curves.

A limitation of calibration curves is that they evaluate only global calibration over the entire test set, without accounting for local calibration, e.g., differences in calibration for specific groups of predictions. To address this, we employ a method similar to the reliability diagrams proposed by Levi et al.\cite{levi2022evaluating}. Predictions are grouped based on their uncertainty values into $n_{\text{bins}}$ bins with intervals
\begin{align}
    I_\beta = \left[\left(\beta - 1\right)\Delta u,\, \beta\Delta u\right),\qquad\text{with}\qquad \beta=1,...,n_\text{bins}.
\end{align}
Different to Levi et al., we use equidistant bins, with a fixed bin size $\Delta u$, and assume that all predictions inside one bin approximately have the center of that bin
\begin{align}
    u_\beta(\mathbf{x}) = \left( \beta - \frac{1}{2} \right) \Delta u
\end{align}
as same fixed uncertainty. 

\begin{figure}
    \centering
    \includegraphics[width=0.5\linewidth]{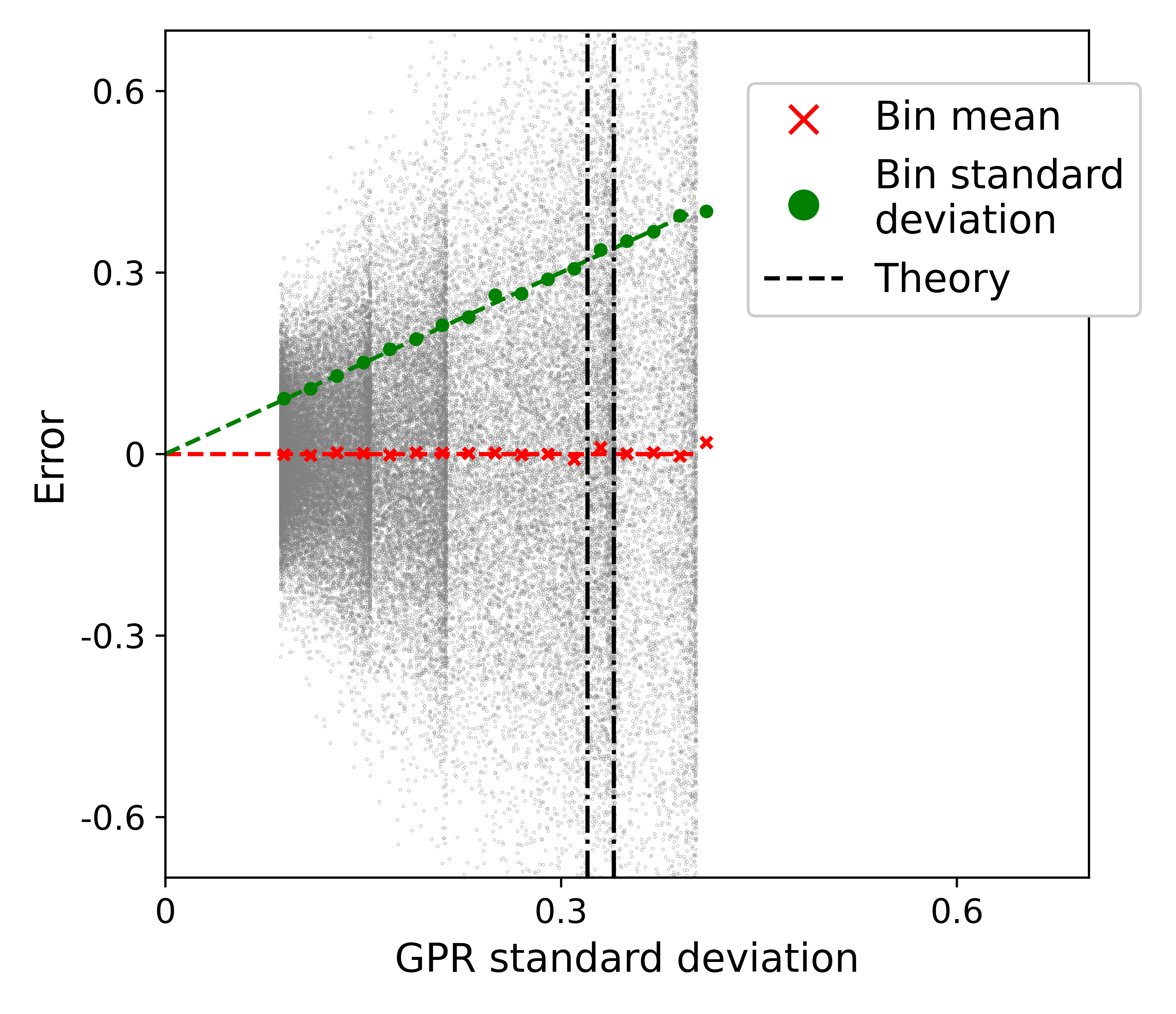}
    \caption{Extended reliability diagram for sinus data with synthetic GPR test targets. To generate the data, a GPR model was trained with noisy data points from a sinus function. Test targets where drawn from the predictive distributions of the GPR model. Thereby data is generated for which the GPR model perfectly captures the underlying data-generating process. For all test samples, the predicted GPR standard deviation is plotted against the actual error of the prediction.  All predictions are separated by their standard deviation into equidistant bins. Mean and standard deviation of error distributions are shown for every bin.}
    \label{fig:synthetic_gpr}
\end{figure}

To evaluate whether the true error distributions align with the model's predictions, we calculate the \textit{bin standard deviation} of the errors $\epsilon_\beta$ within each bin and compare it with $u_\beta\left(\textbf{x}\right)$, which is the predicted standard deviation according to eq.~\eqref{eq:error_dist}. This evaluates whether the errors are as broad or narrow as predicted, but not if they are systematically shifted. In contrast to Levi et al., we therefore also calculate the \textit{bin mean} of the errors $\epsilon_\beta$ within each bin. If the error mean differs from zero, this indicates a systematical shift of errors and suggests that the predictive distribution is shifted with a constant bias compared to the true distribution of targets. We refer to plots where uncertainty estimations are plotted against the errors, with bin means and standard deviations of errors marked, as \textit{extended reliability diagrams}. To illustrate a model perfectly aligned with eq.~\eqref{eq:error_dist}, Fig.~\ref{fig:synthetic_gpr} shows an extended reliability diagrams for data where the targets were drawn from the predictive distribution of the GPR model. 

\subsection{Uncertainty sampling}
Beside explicitly evaluating calibration, we evaluate uncertainty estimations when utilized in an AL scheme, namely uncertainty sampling \cite{uncertainty_sampling}, to see, how effective these estimations are in practical use. In uncertainty sampling, a small random subset of a fixed dataset is selected, and labels are generated for this subset. In our case, these labels are obtained via quantum chemical calculations of potential energies. This labeled data is then used to train an initial model. The uncertainty estimation of this model is used to iteratively select samples from the remaining pool of unlabeled samples for which the labels are generated and which are then added as additional training samples. See Algorithm \ref{alg:uncertainty_sampling} for pseudocode.

To analyze, how much uncertainty sampling with different uncertainty measures increases data efficiency, we kept a test set separate from the data pool. In every iteration, we calculated the mean absolute error (MAE) for the test samples. Additionally, in every iteration, we calculated the maximum absolute error and the variance of the absolute errors of the test samples, to understand how consistently the model is performing across different data points. We compare adding samples via uncertainty sampling with adding samples randomly. We expect the highest increase in data efficiency, if we would have an uncertainty estimation that estimates the true error correctly. Therefore, to understand how much data efficiency could be achieved at maximum, we also iteratively added the sample with the largest absolute error in each iteration.

If, instead of adding one sample per iteration, batches of samples are added, there is a risk of adding multiple very similar samples, for all of whom the model has a high uncertainty. Although methods exist that balance informativeness, diversity, and representativeness\cite{zaverkin2022exploring, wilson2022batch}, we do not use batch active learning approaches.
Our goal to isolate the influence of the uncertainty estimation and get an accurate picture of how much additional information is extracted with different uncertainty estimations.

\begin{algorithm}[H]
\small
\caption{Uncertainty Sampling}
\begin{algorithmic}[1]
\REQUIRE Unlabeled data pool $U_\text{pool}$, initial sample size $n_\text{init}$, number of iterations $n_\text{iter}$
\STATE Randomly select $n_\text{init}$ samples from $U_\text{pool}$ to get $U_\text{init}$ and obtain their labels to form initial labeled dataset $L$
\STATE $U_\text{pool}$ = $U_\text{pool}$ \textbackslash $U_\text{init}$
\STATE Train initial model $m$ using labeled data $L$
\FOR{$i = 1$ to $n_\text{iter}$}
    \STATE Compute uncertainty for each sample in $U_\text{pool}$ using model $m$
    \STATE Select sample $x^*$ from $U_\text{pool}$ with highest uncertainty
    \STATE Obtain label $y^*$ for selected sample $x^*$
    \STATE $L = L \,\cup\, \{(x^*, y^*)\}$
    \STATE $U_\text{pool} = U_\text{pool}$ \textbackslash $\{x^*\}$
    \STATE Retrain model $m$ using updated labeled dataset $L$
\ENDFOR
\STATE \textbf{return} Trained model $m$
\end{algorithmic}
\label{alg:uncertainty_sampling}
\end{algorithm}

\subsection{Data}
MD17\cite{md17} is a benchmark dataset of long molecular dynamic trajectories, calculated with density functional theory (DFT), for 10 small molecules.
Christensen et al.\cite{rmd17} recalculated the energies as the original values were found to be noisy. Additionally, the total number of snapshots from the separate trajectory was reduced to a selection of less correlated snapshots. These changes result in the revised MD17 (rMD17) dataset. We selected the rMD17 dataset for our tests because it is a well-established benchmark, allowing for easy comparison and categorization of our results. We performed tests for potential energies of Benzene, as a highly symmetric molecule and Aspirin, as a more complex molecule.

The WS22 dataset\cite{ws22} consists of distributions of configurations obtained via Wigner sampling for 10 different organic molecules, which aims to cover an extensive configuration space. This is important for our tests to understand how uncertainty estimations can distinguish between areas of configuration spaces where predictions are accurate, and regions where predictions are poor across a wide range of configurations.
We performed tests for potential energies of 2-(methyliminomethyl)phenol (SMA) and 4-(2-hydroxybenzylidene)-1,2-dimethyl-1H-imidazol-5(4H)-one (O-HBDI).

\begin{figure}[ht]
	\centering \includegraphics[width=\textwidth]{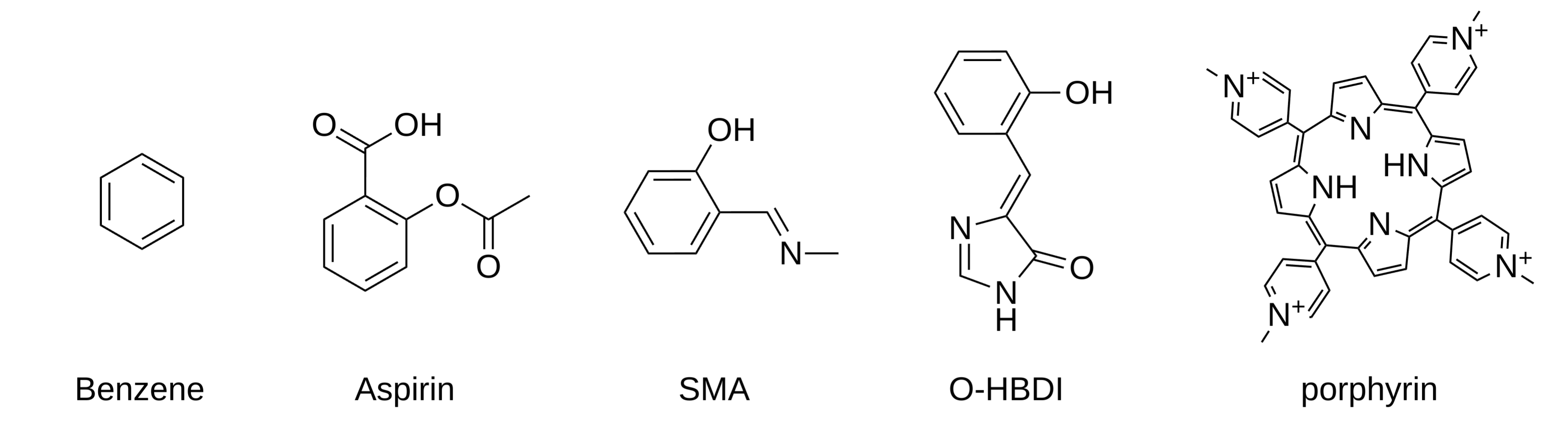}
	\caption{\label{fig:molecules} Chemical structures of the molecules used in the tests.}
 \end{figure}

Additionally, we considered porphyrin as a large and complex molecule with a potentially complex configuration space. The dataset consists of 100,000 snapshots of a quantum molecular dynamic trajectory calculated with Density Functional Tight Binding (DFTB). To calculate the DFTB trajectory we took the initial structure from ChemSpider\cite{chemspider} (ID: 4086). After DFT (B3LYP/6-31G$^{\ast}$) geometry optimization, 150,000 fs ground state dynamics at 300 K were performed with the DFTB+ package\cite{dftb+}, and 100,000 frames were extracted from the end of the trajectory with a stride of 1 fs. We performed our tests for the corresponding excitation energies, which we calculated with time-dependent long-range corrected DFTB (TD-LC-DFTB). Figure~\ref{fig:molecules} gives the chemical structures of all molecules used in this study.

\section{Results}\label{sec:results}
For our experiments, we randomly select 1000 training samples and 2000 test samples from every dataset, and the remaining samples form a candidate pool. For every dataset, we use the 1000 training samples to optimize the hyperparameters. For the calibration curves and extended reliability diagrams, we calculate the uncertainty estimations and prediction errors on the candidate pool, with the respective GPR models trained on the 1000 training samples. In the extended reliability diagrams we separate the samples into equidistant bins by their uncertainty estimations and calculate the bin means and standard deviations of the errors. For the different datasets, we manually select the bin width based on the range of predicted uncertainties. The exact bin widths can be found in table S1 of the supplementary material. We compare the calculated bin means and standard deviation with the theoretical values of Eq.~\ref{eq:error_dist}. We only show the bin mean and standard deviation, if the number of samples inside one bin is greater than 50. For better visibility, we manually adjust the axis limits of the extended reliability diagrams, therefore a few scatter points are not displayed. For all uncertainty sampling tests, we use the fixed optimized hyperparameters and the test set, as described above. For the initial model, only 200 training samples are used, randomly selected from the dataset excluding the 2,000 test samples.

\subsection{Poor global calibration for ensemble uncertainties}\label{sec:results_diff_uncertainty_measures}

\begin{figure}[h!]
    \centering
    \begin{subfigure}[b]{\textwidth}
        \centering
        \includegraphics[width=0.8\textwidth]{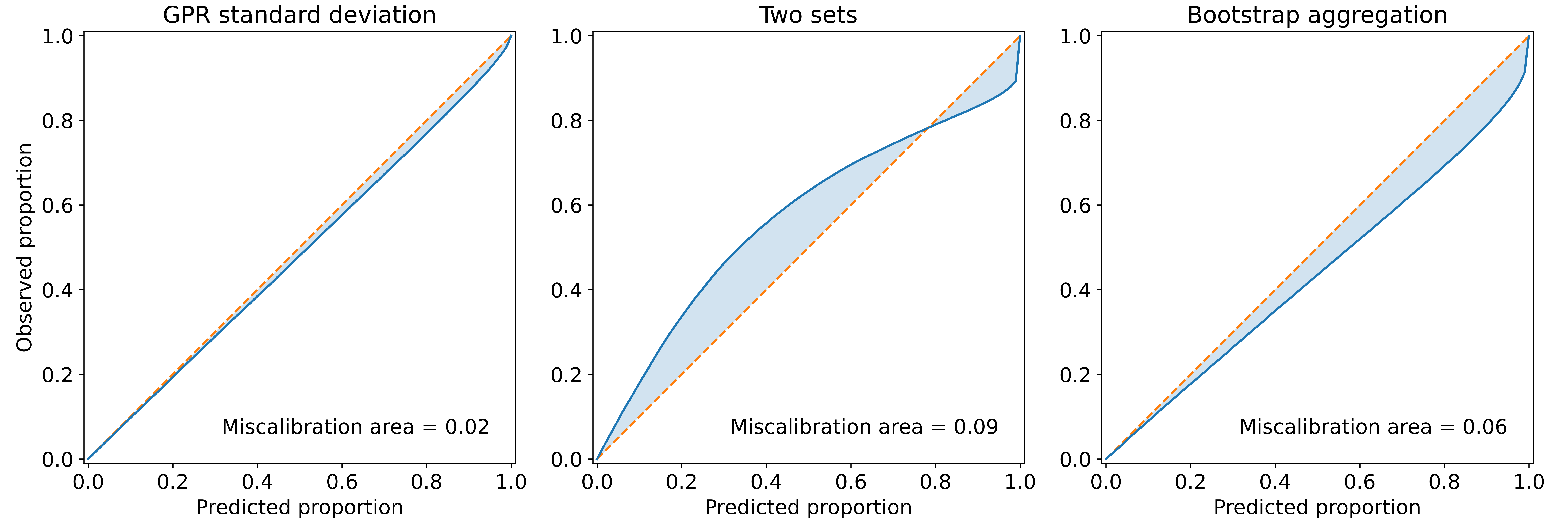}
        \caption{Benzene}
    \end{subfigure}
    \\
    \begin{subfigure}[b]{\textwidth}
        \centering
        \includegraphics[width=0.8\textwidth]{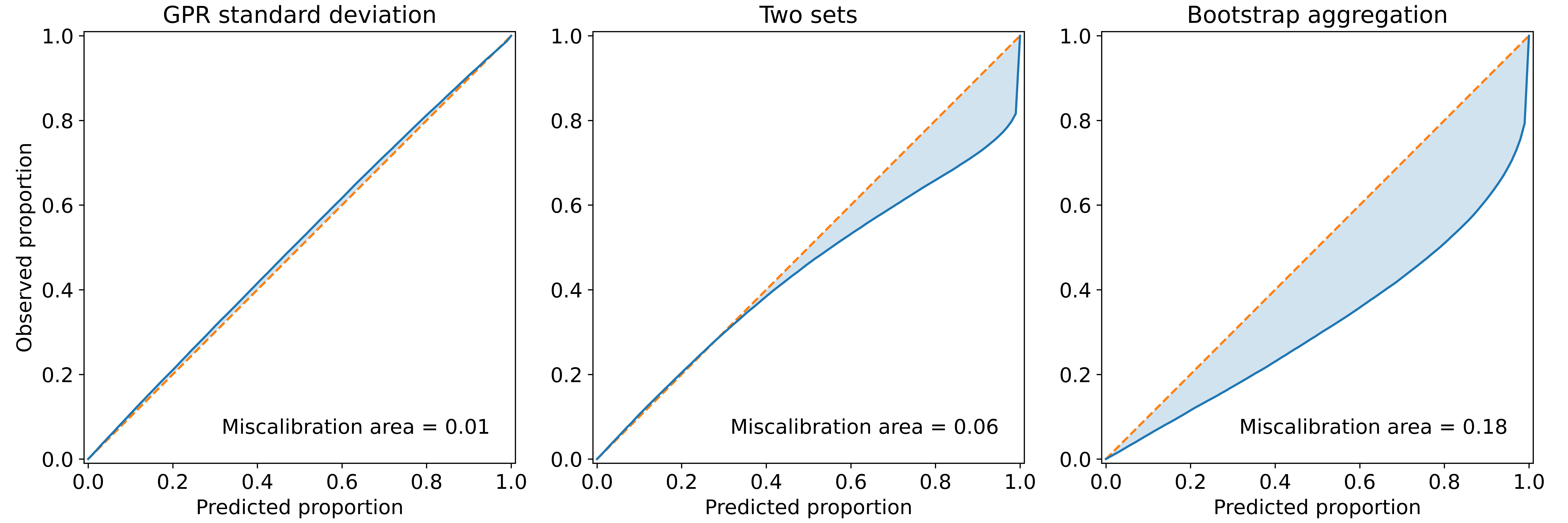}
        \caption{SMA}
    \end{subfigure}
    \caption{Calibration curves of different uncertainty measures of GPR with SOAP for rMD17 benzene and WS22 SMA. Calibration curves plot the predicted proportion of test data expected to fall within $\alpha$-prediction intervals (horizontal axis) against the observed proportion of test data that actually falls within these intervals (vertical axis), iterating over different values of $\alpha$.}
    \label{fig:compare_uncertainty_measure_calibration_soap}
\end{figure}

In Fig.~\ref{fig:compare_uncertainty_measure_calibration_soap}, calibration curves for all three uncertainty measures are shown. We exemplary show calibration curves of GPR with SOAP for rMD17 benzene and WS22 SMA. The calibration curve for the GPR standard deviation is nearly diagonal and has a low miscalibration area, indicating relatively good calibration across the dataset. In contrast, the two sets and bootstrap uncertainty measures show much poorer calibration, with significantly higher miscalibration areas. In Fig.~\ref{fig:coulomb_GPR_model} and in Figs.~S3-7 of the supplementary material calibration curves for all uncertainty measures across all molecules are shown for GPR with Coulomb as well as GPR with SOAP. In nearly all cases, the GPR standard deviation demonstrates much better global calibration, with notably lower miscalibration areas compared to the other two uncertainty measures. The only exception is GPR with Coulomb for benzene. Here, the calibration curve for the GPR standard deviation shows a relatively high miscalibration area, which exceeds that of the two sets uncertainty measure.

\subsection{Local calibration analysis reveals model bias}
\begin{figure}
    \centering
    \includegraphics[width=\textwidth]{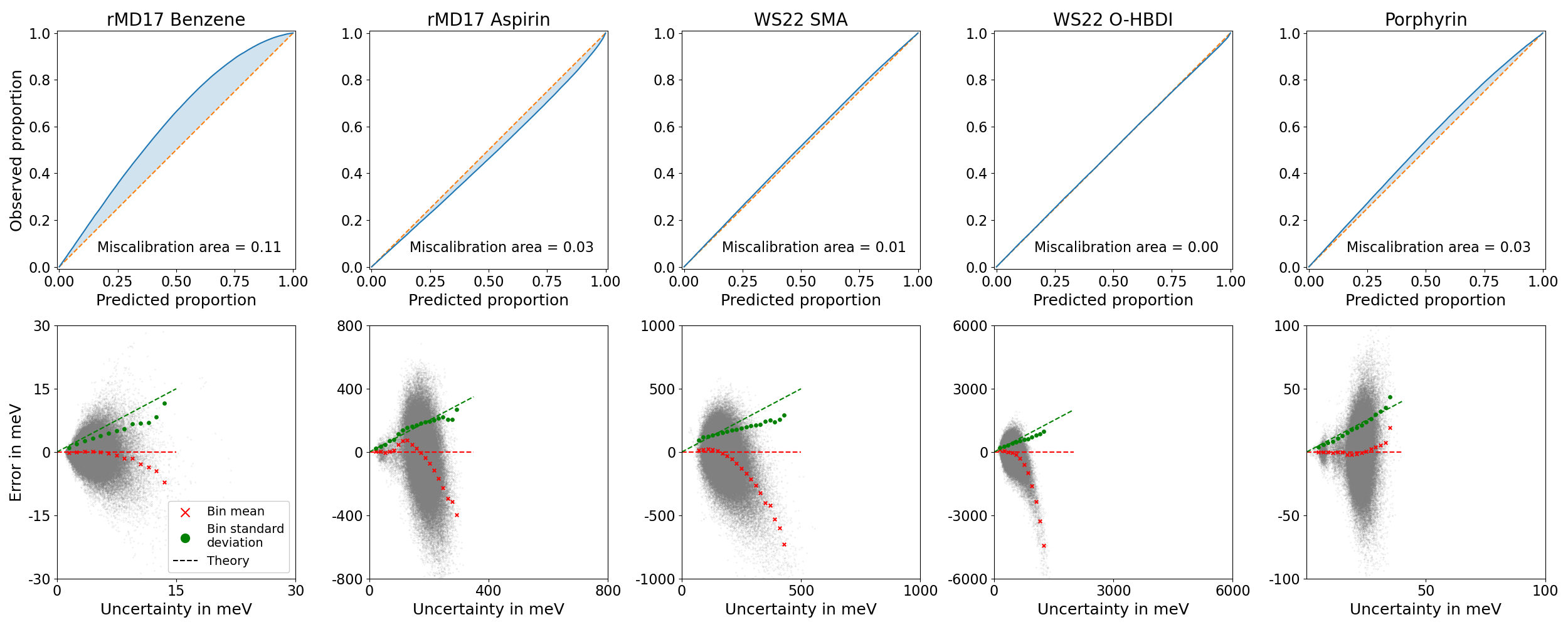}
    \caption{Calibration curves and extended reliability diagrams of the GPR standard deviation of GPR with Coulomb applied to different datasets.  For the extended reliability diagrams, samples were separated into equidistant bins by their uncertainty, and the mean and standard deviation of the errors of all samples in one bin were calculated and are compared with the theoretical values from Eq.~\ref{eq:error_dist}.}
    \label{fig:coulomb_GPR_model}
\end{figure}

While the calibration curves indicate good global calibration for the GPR standard deviation, this does not mean that the model provides a well-calibrated uncertainty estimation for every prediction. Fig.~\ref{fig:coulomb_GPR_model} presents calibration curves together with extended reliability diagrams for GPR with Coulomb across all molecules for the GPR standard deviation. As described above, the calibration curves indicate good global calibration for all molecules except benzene. However, when predictions are grouped by their uncertainty, the extended reliability diagrams reveal poor calibration for specific groups of predictions. While the standard deviation of errors often aligns rather closely with the theoretical value, the mean value of errors can deviate strongly from zero, particularly for predictions with high uncertainties. Assuming that the errors within separate bins, for which the means and standard deviations are calculated, are distributed somewhat close to normal, this can be interpreted as a systematic shift in the error distributions. In Fig.~S8 of the supplementary material, we explicitly show the error distributions within selected bins for GPR with Coulomb applied to Aspirin and O-HBDI. Indeed, these errors rather closely follow normal distributions hat are shifted relative to the predicted normal distribution. This systematic shift leads to much higher errors than predicted and indicates a systematic bias in the respective predictions. Although this bias is not captured quantitatively by the uncertainty estimation, it can still be observed that, in most cases, there is a clear correlation between increasing GPR standard deviation and increasing bias. This means the GPR standard deviation can identify predictions with a high bias. 

In ML theory, the error of a model is often divided into an irreducible part resulting from the noise in the training data, as well as a bias and a variance component. Hüllermeier and Waegeman\cite{hullermeier2021aleatoric} describe the GPR standard deviation as a meaningful indicator of the total model uncertainty. However, if we have a model that has a bias in the form of systematical shifts of the predictive distributions in certain areas of the input space, it is not possible for the standard deviation of the predictive distribution to quantify these shifts. In general, a bias is associated with incorrect model assumptions, making it challenging to quantify the resulting error using the flawed model itself. For ensemble-based uncertainties, Heid et al.\cite{heid2023characterizing} discussed that a bias cannot be quantified.

Extended reliability diagrams for the GPR standard deviation of GPR with SOAP are shown in Fig.~S5 of the supplementary material. 
In general, similar trends can be observed, although the bias is lower, and in some cases, the standard deviation of errors deviates more strongly from the theoretical value. Extended reliability diagrams for the two sets and bootstrap uncertainty of GPR wit Coulomb as well as GPR with SOAP can be found in Figs.~S3-4 and S6-7 of the supplementary material. For the two sets uncertainty, it can be observed in all cases that errors are distributed with almost identical standard deviations and means across all bins. Consequently, the reliability diagrams reveal that the two set uncertainty not only shows poor global calibration but also provides no meaningful information to differentiate between predictions with different errors. A similar observation can be made for the bootstrap uncertainty in some of the cases.

\subsection{Uncertainty sampling results in worse error than random sampling}
\begin{figure}[h!]
    \centering
    \begin{subfigure}[b]{\textwidth}
        \includegraphics[width=\textwidth]{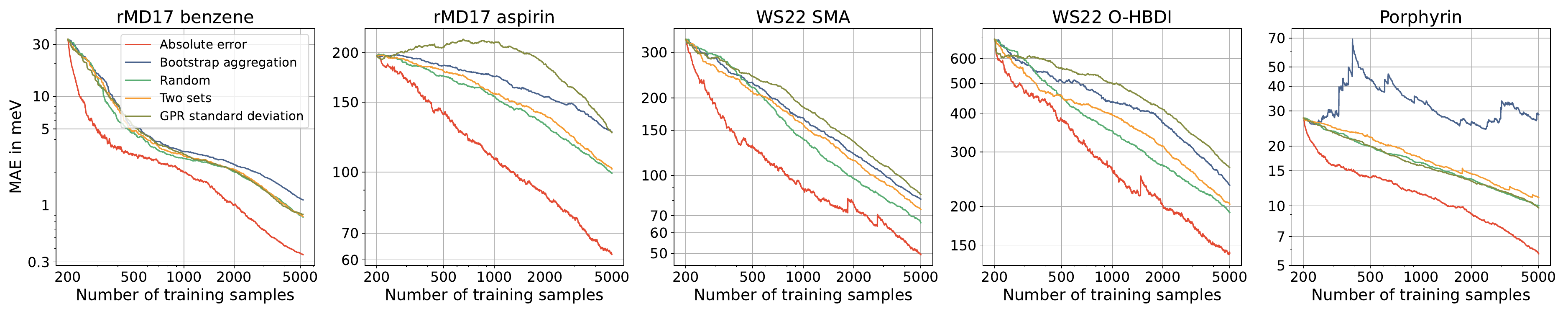}
        \caption{GPR with Coulomb}
        \label{fig:cbal_diff_datasets_coulomb}
    \end{subfigure}
    \\
    \begin{subfigure}[b]{\textwidth}
        \centering
        \includegraphics[width=0.8\textwidth]{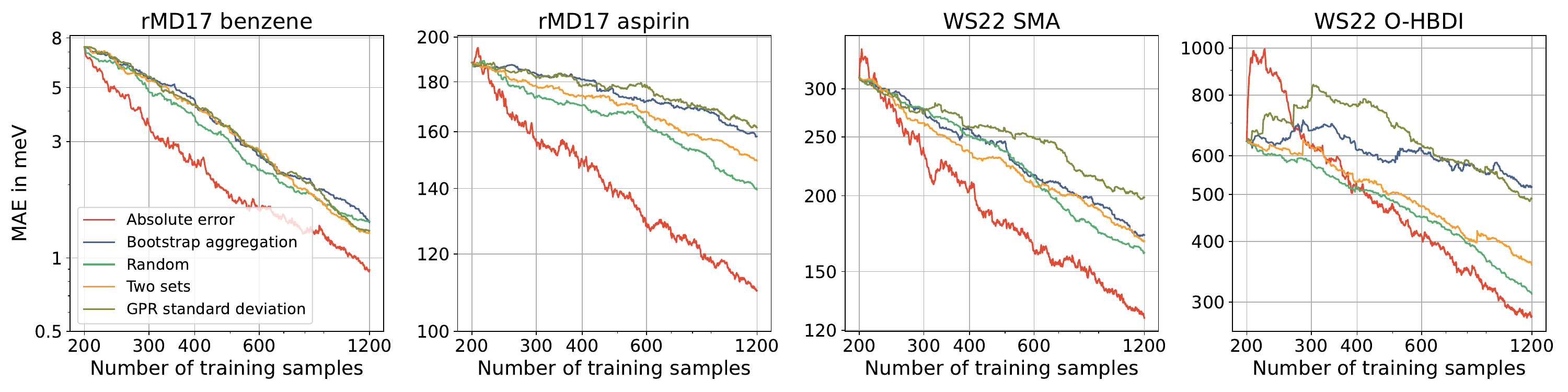}
        \caption{GPR with SOAP}
        \label{fig:cbal_diff_datasets_soap}
    \end{subfigure}
    \caption{Uncertainty sampling results for GPR with Coulomb and SOAP representations for different datasets. We randomly selected 200 samples as initial training samples and 2000 samples as test samples. From the remaining pool of samples, in every iteration, the sample with the highest respective uncertainty was selected and added as an additional training sample. Additionally we randomly added a new sample in every iteration and we added the sample with the highest actual absolute error. During each iteration, we compute the MAE.}
    \label{fig:uncertainty_sampling_mae}
\end{figure}

In Fig.~\ref{fig:uncertainty_sampling_mae}, the MAE for every iteration of uncertainty sampling across all molecules is shown. In most cases, uncertainty sampling with the GPR standard deviation results in a worse MAE compared to adding samples randomly. Only for benzene and porphyrin with Coulomb matrices, uncertainty sampling results in a similar MAE compared to adding samples randomly. Note that in the local calibration analysis, this were exactly the cases where the extended reliability diagrams did not show a significant bias for predictions with high uncertainties. The two sets uncertainty performs very similar to random sampling, as expected, since the extended reliability diagrams demonstrate that it does not provide any meaningful information about the error. Interestingly, uncertainty sampling with the bootstrap aggregation uncertainty results in a worse MAE than random sampling in almost all cases. When samples corresponding to predictions with the highest absolute error were added to the training set, the MAE improved significantly for all cases, showing that in principle there is a relevant potential for an improved data efficiency through better sampling strategies.

Our findings align well with those of Uteva et al.\cite{uteva2018active}, who reported that uncertainty sampling with the GPR variance performs worse than uncertainty sampling with the two sets uncertainty, adding samples corresponding to predictions with the highest absolute error, and training with maximum Latin hypercube data. The reduced data efficiency observed when sampling a fixed configuration space using uncertainty sampling suggests that, in on-the-fly active learning (AL), where uncertainty estimations guide the exploration of the configuration space, data efficiency may also be compromised. In contrast, Vandermause et al.\cite{flare} demonstrated increased data efficiency with their on-the-fly AL scheme applied to five-element high-entropy alloy structures.

Even if the MAE is not improved by uncertainty sampling, a model trained via uncertainty sampling may perform more consistently across different data points in the test set. To analyze this, we present results for the maximum absolute test error and the test variance in Figs.~S9 and S10 of the supplementary material. Indeed, the maximum absolute error of the test set improves for uncertainty sampling with the GPR standard deviation in all cases, except for aspirin. The test variance, however, is similar to that of random sampling in most cases or deteriorates, with improvements observed only for benzene and O-HBDI with Coulomb. When training samples corresponding to predictions with the highest absolute error were added, both metrics improved significantly for all cases.

\subsection{More general model after uncertainty sampling?}
\begin{figure}[h!]
    \centering
    \begin{subfigure}[b]{\textwidth}
        \centering
        \includegraphics[width=0.8\textwidth]{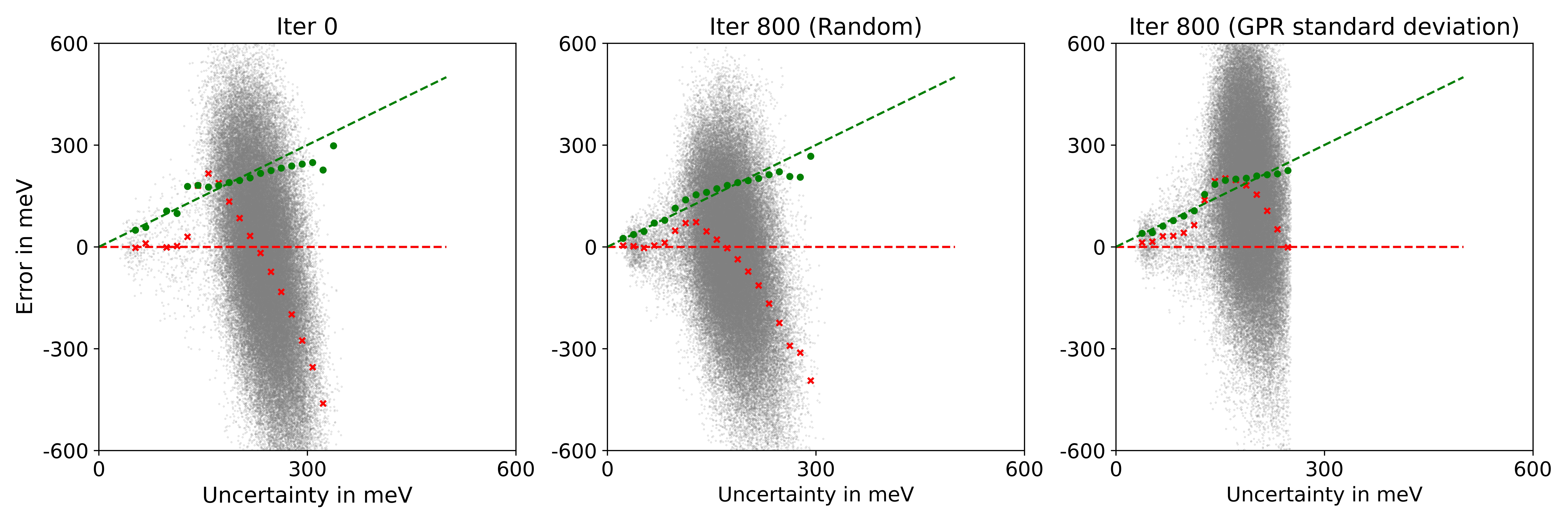}
        \caption{Aspirin}
        \label{fig:us_vs_calibration_aspirin}
    \end{subfigure}
    \\
    \begin{subfigure}[b]{\textwidth}
        \centering
        \includegraphics[width=0.8\textwidth]{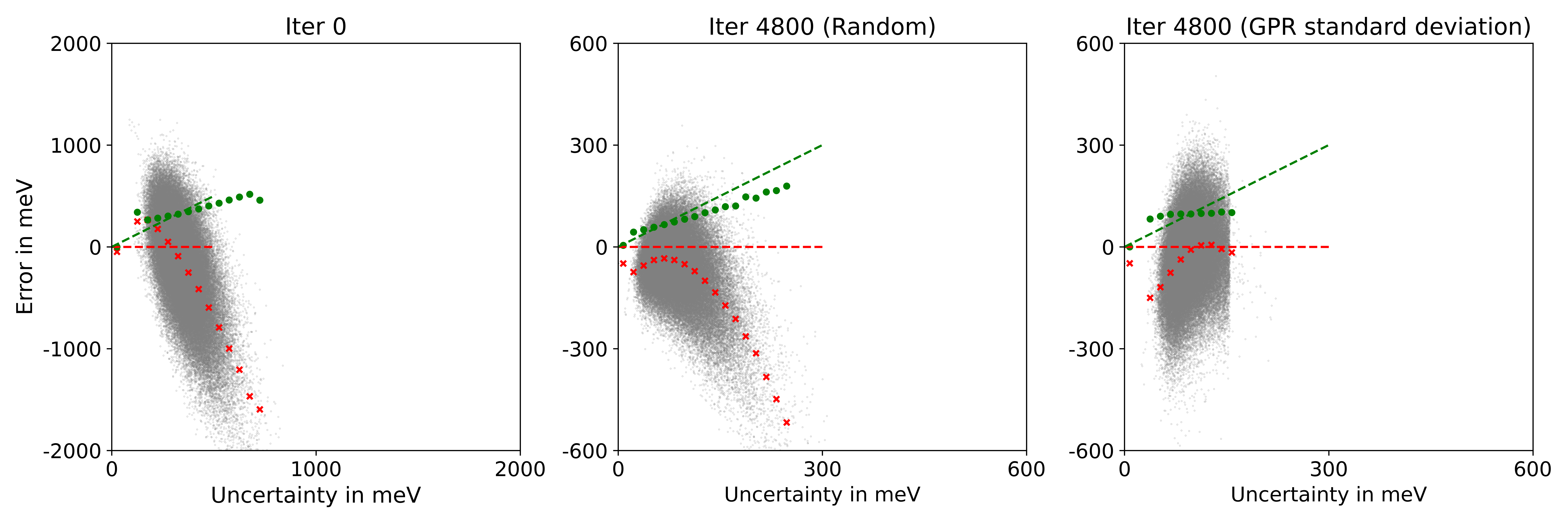}
        \caption{SMA}
        \label{fig:us_vs_calibration_sma}
    \end{subfigure}
    \caption{Extended reliability diagrams of GPR with Coulomb representations for aspirin and SMA are presented. These plots compare the initial GPR models trained on 200 random configurations with those obtained after uncertainty sampling using the GPR standard deviation, as well as models updated by adding samples randomly. For better visibility, the plots for SMA use different bin sizes and axis limits between the initial model and those following uncertainty sampling. Note that the axis values also differ from those in Fig.~\ref{fig:coulomb_GPR_model}.}
    \label{fig:us_vs_calibration}
\end{figure}

The significantly worse MAE for uncertainty sampling with the GPR standard deviation compared to random sampling might seem surprising, as we observe a correlation between increasing GPR standard deviation and increasing bias. This suggests that adding configurations corresponding to the highest GPR standard deviation as training samples should be beneficial. To analyze this behavior, in Fig.~\ref{fig:us_vs_calibration} we show extended reliability diagrams after uncertainty sampling with the GPR standard deviation and after random sampling for GPR with Coulomb applied to Aspirin and SMA. For SMA, we show the plots for the models after all 4800 iterations, while for Aspirin, the plots are shown after only 800 iterations, because at that point, the model trained via uncertainty sampling with the GPR standard deviation performs even worse than the initial model. Additionally, we show the plot for the initial GPR model trained with 200 random samples. In both cases, it can be observed that the initial model not only has a high bias for predictions with high uncertainties but also a smaller bias in the opposite direction for predictions with very low uncertainty. After uncertainty sampling, the high bias for predictions with high uncertainty is eliminated in both cases. However, the bias for predictions with low uncertainty is higher than after random sampling and, for Aspirin, even higher than for the initial model. The standard deviation of errors of the respective bins with low uncertainty is higher than after random sampling too. 

These observations could be explained by the fact that uncertainty sampling with the GPR standard deviation finds a maximally distant training set. This may result in a model that strongly emphasizes the borders of the configuration space originally sampled in the fixed dataset. In such fixed datasets, the borders of the configuration space might be sampled more sparsely, while other areas are sampled more densely. A model that emphasizes the borders strongly might perform slightly worse on samples from other areas of the configuration space, which are more frequent. While this means that uncertainty sampling can result in an increased average model error within the fixed configuration space, it is likely to be beneficial for extrapolation tasks. This observation was also made by Podryabinkin et al.\cite{podryabinkin2017active}, who trained an MLIP using the most uncertain configurations of a fixed dataset, based on the D-optimality criterion. Compared to random sampling, the model shows a slightly increased root mean squared error and a slightly decreased maximum error when tested on a dataset randomly selected from the fixed dataset. However, MD simulations using this model were significantly more stable.

\section{Conclusion}\label{sec:conclusions}
In this study, we evaluated calibration of different uncertainty measures for GPR based MLIPs and their performance when applied to active learning, where we iteratively choose the configuration with the maximum uncertainty from a fixed configuration space. Regarding calibration, it becomes evident how important it is to consider not only global calibration, which averages over the entire test dataset, but also local calibration, which considers specific groups of predictions. Ensemble-based uncertainty measures appear to be a poor choice for GPR-based MLIPs, as they already show poor global calibration. In contrast, the GPR standard deviation demonstrates relatively good global calibration. However, when calibration is evaluated locally, the GPR standard deviation can show very poor calibration. We find that, in particular, groups of predictions with a high GPR standard deviation often have a high bias that is not captured quantitatively by the uncertainty estimation.
Therefore, without further knowledge, the GPR standard deviation should not be interpreted as a quantitative measure of potential error ranges, such as confidence intervals derived from the GPR standard deviation. Still, we find that an increasing GPR standard deviation correlates with an increasing bias and can therefore identify predictions with high bias and high error. When sampling a fixed configuration space by selecting configurations for which a GPR model predicts the maximum GPR standard deviation, the resulting training configurations are maximally distant from each other. This may lead to a more general model, that performs better in extrapolation tasks. 
However, such a model may overemphasize the borders of the configuration space in a fixed dataset, resulting in worse average performance on the fixed dataset compared to a model trained using random sampling.


\paragraph{Acknowledgment}
The authors acknowledge support by the DFG through the DFG Priority Program SPP 2363 on “Utilization and Development of Machine Learning for Molecular Applications – Molecular Machine Learning” through the projects ZA 1175/4-1 and KL 1299/25-1.  Some of the computations were carried out on the \href{https://pleiades.uni-wuppertal.de}{PLEIADES cluster} at the University of Wuppertal, which was supported by the Deutsche Forschungsgemeinschaft (DFG, grant No. INST 218/78-1 FUGG) and the Bundesministerium für Bildung und Forschung (BMBF).
The authors would also like to acknowledge the support of the `Interdisciplinary Center for Machine Learning and Data Analytics (IZMD)' at the University of Wuppertal. 
The authors also acknowledge the use of ChatGPT-4o, for assistance with language refinement, grammar corrections, and reformulations in the preparation of this manuscript.
\paragraph{Data availability statement}
The data that support the findings of this study are openly available at the following URL: xxx

\section*{References}
\bibliography{literature}

\begin{thebibliography}{10}

\bibitem{rupp2014machine}
Matthias Rupp, Matthias~R Bauer, Rainer Wilcken, Andreas Lange, Michael Reutlinger, Frank~M Boeckler, and Gisbert Schneider.
\newblock Machine learning estimates of natural product conformational energies.
\newblock {\em PLoS computational biology}, 10(1):e1003400, 2014.

\bibitem{chen2023accelerating}
Benjamin~WJ Chen, Xinglong Zhang, and Jia Zhang.
\newblock Accelerating explicit solvent models of heterogeneous catalysts with machine learning interatomic potentials.
\newblock {\em Chemical Science}, 14(31):8338--8354, 2023.

\bibitem{flare}
Jonathan Vandermause, Steven~B Torrisi, Simon Batzner, Yu~Xie, Lixin Sun, Alexie~M Kolpak, and Boris Kozinsky.
\newblock On-the-fly active learning of interpretable {Bayesian} force fields for atomistic rare events.
\newblock {\em npj Computational Materials}, 6(1):20, 2020.

\bibitem{van2023hyperactive}
Cas van~der Oord, Matthias Sachs, D{\'a}vid~P{\'e}ter Kov{\'a}cs, Christoph Ortner, and G{\'a}bor Cs{\'a}nyi.
\newblock Hyperactive learning for data-driven interatomic potentials.
\newblock {\em npj Computational Materials}, 9(1):168, 2023.

\bibitem{montes2022training}
David Montes~de Oca~Zapiain, Mitchell~A Wood, Nicholas Lubbers, Carlos~Z Pereyra, Aidan~P Thompson, and Danny Perez.
\newblock Training data selection for accuracy and transferability of interatomic potentials.
\newblock {\em npj Computational Materials}, 8(1):189, 2022.

\bibitem{kulichenko2023uncertainty}
Maksim Kulichenko, Kipton Barros, Nicholas Lubbers, Ying~Wai Li, Richard Messerly, Sergei Tretiak, Justin~S Smith, and Benjamin Nebgen.
\newblock Uncertainty-driven dynamics for active learning of interatomic potentials.
\newblock {\em Nature Computational Science}, 3(3):230--239, 2023.

\bibitem{podryabinkin2017active}
Evgeny~V Podryabinkin and Alexander~V Shapeev.
\newblock Active learning of linearly parametrized interatomic potentials.
\newblock {\em Computational Materials Science}, 140:171--180, 2017.

\bibitem{zhang2019active}
Linfeng Zhang, De-Ye Lin, Han Wang, Roberto Car, and E~Weinan.
\newblock Active learning of uniformly accurate interatomic potentials for materials simulation.
\newblock {\em Physical Review Materials}, 3(2):023804, 2019.

\bibitem{uteva2018active}
Elena Uteva, Richard~S Graham, Richard~D Wilkinson, and Richard~J Wheatley.
\newblock Active learning in gaussian process interpolation of potential energy surfaces.
\newblock {\em The Journal of chemical physics}, 149(17), 2018.

\bibitem{zhu2023fast}
Albert Zhu, Simon Batzner, Albert Musaelian, and Boris Kozinsky.
\newblock Fast uncertainty estimates in deep learning interatomic potentials.
\newblock {\em The Journal of Chemical Physics}, 158(16), 2023.

\bibitem{zaverkin2022exploring}
Viktor Zaverkin, David Holzm{\"u}ller, Ingo Steinwart, and Johannes K{\"a}stner.
\newblock Exploring chemical and conformational spaces by batch mode deep active learning.
\newblock {\em Digital Discovery}, 1(5):605--620, 2022.

\bibitem{wilson2022batch}
Nathan Wilson, Daniel Willhelm, Xiaoning Qian, Raymundo Arr{\'o}yave, and Xiaofeng Qian.
\newblock Batch active learning for accelerating the development of interatomic potentials.
\newblock {\em Computational Materials Science}, 208:111330, 2022.

\bibitem{toyoura2016machine}
Kazuaki Toyoura, Daisuke Hirano, Atsuto Seko, Motoki Shiga, Akihide Kuwabara, Masayuki Karasuyama, Kazuki Shitara, and Ichiro Takeuchi.
\newblock Machine-learning-based selective sampling procedure for identifying the low-energy region in a potential energy surface: A case study on proton conduction in oxides.
\newblock {\em Physical Review B}, 93(5):054112, 2016.

\bibitem{sivaraman2020machine}
Ganesh Sivaraman, Anand~Narayanan Krishnamoorthy, Matthias Baur, Christian Holm, Marius Stan, G{\'a}bor Cs{\'a}nyi, Chris Benmore, and {\'A}lvaro V{\'a}zquez-Mayagoitia.
\newblock Machine-learned interatomic potentials by active learning: amorphous and liquid hafnium dioxide.
\newblock {\em npj Computational Materials}, 6(1):104, 2020.

\bibitem{lin2020automatically}
Qidong Lin, Yaolong Zhang, Bin Zhao, and Bin Jiang.
\newblock Automatically growing global reactive neural network potential energy surfaces: A trajectory-free active learning strategy.
\newblock {\em The Journal of Chemical Physics}, 152(15), 2020.

\bibitem{guan2018construction}
Yafu Guan, Shuo Yang, and Dong~H Zhang.
\newblock Construction of reactive potential energy surfaces with {Gaussian} process regression: active data selection.
\newblock {\em Molecular Physics}, 116(7-8):823--834, 2018.

\bibitem{schwalbe2021differentiable}
Daniel Schwalbe-Koda, Aik~Rui Tan, and Rafael G{\'o}mez-Bombarelli.
\newblock Differentiable sampling of molecular geometries with uncertainty-based adversarial attacks.
\newblock {\em Nature communications}, 12(1):5104, 2021.

\bibitem{schnet}
Kristof~T Sch{\"u}tt, Huziel~E Sauceda, P-J Kindermans, Alexandre Tkatchenko, and K-R M{\"u}ller.
\newblock Schnet--a deep learning architecture for molecules and materials.
\newblock {\em The Journal of Chemical Physics}, 148(24), 2018.

\bibitem{painn}
Kristof Sch{\"u}tt, Oliver Unke, and Michael Gastegger.
\newblock Equivariant message passing for the prediction of tensorial properties and molecular spectra.
\newblock In {\em International Conference on Machine Learning}, pages 9377--9388. PMLR, 2021.

\bibitem{nequip}
Simon Batzner, Albert Musaelian, Lixin Sun, Mario Geiger, Jonathan~P Mailoa, Mordechai Kornbluth, Nicola Molinari, Tess~E Smidt, and Boris Kozinsky.
\newblock E(3)-equivariant graph neural networks for data-efficient and accurate interatomic potentials.
\newblock {\em Nature communications}, 13(1):2453, 2022.

\bibitem{mace}
Ilyes Batatia, David~P Kovacs, Gregor Simm, Christoph Ortner, and G{\'a}bor Cs{\'a}nyi.
\newblock {MACE}: Higher order equivariant message passing neural networks for fast and accurate force fields.
\newblock {\em Advances in Neural Information Processing Systems}, 35:11423--11436, 2022.

\bibitem{coulomb}
Matthias Rupp, Alexandre Tkatchenko, Klaus-Robert M{\"u}ller, and O~Anatole Von~Lilienfeld.
\newblock Fast and accurate modeling of molecular atomization energies with machine learning.
\newblock {\em Physical review letters}, 108(5):058301, 2012.

\bibitem{mbtp}
Haoyan Huo and Matthias Rupp.
\newblock Unified representation of molecules and crystals for machine learning.
\newblock {\em Machine Learning: Science and Technology}, 3(4):045017, 2022.

\bibitem{bag_of_bonds}
Katja Hansen, Franziska Biegler, Raghunathan Ramakrishnan, Wiktor Pronobis, O~Anatole Von~Lilienfeld, Klaus-Robert Muller, and Alexandre Tkatchenko.
\newblock Machine learning predictions of molecular properties: Accurate many-body potentials and nonlocality in chemical space.
\newblock {\em The journal of physical chemistry letters}, 6(12):2326--2331, 2015.

\bibitem{soap}
Albert~P Bart{\'o}k, Risi Kondor, and G{\'a}bor Cs{\'a}nyi.
\newblock On representing chemical environments.
\newblock {\em Physical Review B—Condensed Matter and Materials Physics}, 87(18):184115, 2013.

\bibitem{snap}
Aidan~P Thompson, Laura~P Swiler, Christian~R Trott, Stephen~M Foiles, and Garritt~J Tucker.
\newblock Spectral neighbor analysis method for automated generation of quantum-accurate interatomic potentials.
\newblock {\em Journal of Computational Physics}, 285:316--330, 2015.

\bibitem{ace}
Ralf Drautz.
\newblock Atomic cluster expansion for accurate and transferable interatomic potentials.
\newblock {\em Physical Review B}, 99(1):014104, 2019.

\bibitem{behler_parrinello}
J{\"o}rg Behler and Michele Parrinello.
\newblock Generalized neural-network representation of high-dimensional potential-energy surfaces.
\newblock {\em Physical review letters}, 98(14):146401, 2007.

\bibitem{langer2022representations}
Marcel~F Langer, Alex Goe{\ss}mann, and Matthias Rupp.
\newblock Representations of molecules and materials for interpolation of quantum-mechanical simulations via machine learning.
\newblock {\em npj Computational Materials}, 8(1):41, 2022.

\bibitem{linear_ace}
D{\'a}vid~P{\'e}ter Kov{\'a}cs, Cas van~der Oord, Jiri Kucera, Alice~EA Allen, Daniel~J Cole, Christoph Ortner, and G{\'a}bor Cs{\'a}nyi.
\newblock Linear atomic cluster expansion force fields for organic molecules: beyond {RMSE}.
\newblock {\em Journal of chemical theory and computation}, 17(12):7696--7711, 2021.

\bibitem{gaussian_moments}
Viktor Zaverkin and J~K\"astner.
\newblock Gaussian moments as physically inspired molecular descriptors for accurate and scalable machine learning potentials.
\newblock {\em Journal of Chemical Theory and Computation}, 16(8):5410--5421, 2020.

\bibitem{ani}
Justin~S Smith, Olexandr Isayev, and Adrian~E Roitberg.
\newblock {ANI-1}: an extensible neural network potential with {DFT} accuracy at force field computational cost.
\newblock {\em Chemical science}, 8(4):3192--3203, 2017.

\bibitem{gap}
Albert~P Bart{\'o}k, Mike~C Payne, Risi Kondor, and G{\'a}bor Cs{\'a}nyi.
\newblock Gaussian approximation potentials: The accuracy of quantum mechanics, without the electrons.
\newblock {\em Physical review letters}, 104(13):136403, 2010.

\bibitem{williams2006gaussian}
Christopher~KI Williams and Carl~Edward Rasmussen.
\newblock {\em Gaussian processes for machine learning}, volume~2.
\newblock MIT press Cambridge, MA, 2006.

\bibitem{bootstrap}
Bradley Efron.
\newblock Bootstrap methods: another look at the jackknife.
\newblock In {\em Breakthroughs in statistics: Methodology and distribution}, pages 569--593. Springer, 1992.

\bibitem{carrete2023deep}
Jes{\'u}s Carrete, Hadri{\'a}n Montes-Campos, Ralf Wanzenb{\"o}ck, Esther Heid, and Georg~KH Madsen.
\newblock Deep ensembles vs committees for uncertainty estimation in neural-network force fields: Comparison and application to active learning.
\newblock {\em The Journal of Chemical Physics}, 158(20), 2023.

\bibitem{kuleshov2018accurate}
Volodymyr Kuleshov, Nathan Fenner, and Stefano Ermon.
\newblock Accurate uncertainties for deep learning using calibrated regression.
\newblock In {\em International conference on machine learning}, pages 2796--2804. PMLR, 2018.

\bibitem{pernot2022prediction}
Pascal Pernot.
\newblock Prediction uncertainty validation for computational chemists.
\newblock {\em The Journal of Chemical Physics}, 157(14), 2022.

\bibitem{levi2022evaluating}
Dan Levi, Liran Gispan, Niv Giladi, and Ethan Fetaya.
\newblock Evaluating and calibrating uncertainty prediction in regression tasks.
\newblock {\em Sensors}, 22(15):5540, 2022.

\bibitem{tran2020methods}
Kevin Tran, Willie Neiswanger, Junwoong Yoon, Qingyang Zhang, Eric Xing, and Zachary~W Ulissi.
\newblock Methods for comparing uncertainty quantifications for material property predictions.
\newblock {\em Machine Learning: Science and Technology}, 1(2):025006, 2020.

\bibitem{kellner2024uncertainty}
Matthias Kellner and Michele Ceriotti.
\newblock Uncertainty quantification by direct propagation of shallow ensembles.
\newblock {\em Machine Learning: Science and Technology}, 2024.

\bibitem{peterson2017addressing}
Andrew~A Peterson, Rune Christensen, and Alireza Khorshidi.
\newblock Addressing uncertainty in atomistic machine learning.
\newblock {\em Physical Chemistry Chemical Physics}, 19(18):10978--10985, 2017.

\bibitem{busk2023graph}
Jonas Busk, Mikkel~N Schmidt, Ole Winther, Tejs Vegge, and Peter~Bj{\o}rn J{\o}rgensen.
\newblock Graph neural network interatomic potential ensembles with calibrated aleatoric and epistemic uncertainty on energy and forces.
\newblock {\em Physical Chemistry Chemical Physics}, 25(37):25828--25837, 2023.

\bibitem{uncertainty_sampling}
David~D Lewis.
\newblock A sequential algorithm for training text classifiers: Corrigendum and additional data.
\newblock In {\em Acm Sigir Forum}, volume~29, pages 13--19. ACM New York, NY, USA, 1995.

\bibitem{mussmann2018relationship}
Stephen Mussmann and Percy Liang.
\newblock On the relationship between data efficiency and error for uncertainty sampling.
\newblock In {\em International Conference on Machine Learning}, pages 3674--3682. PMLR, 2018.

\bibitem{bartok2022improved}
Albert~P Bart{\'o}k and James Kermode.
\newblock Improved uncertainty quantification for {Gaussian} process regression based interatomic potentials.
\newblock {\em arXiv preprint arXiv:2206.08744}, 2022.

\bibitem{gpytorch}
Jacob Gardner, Geoff Pleiss, Kilian~Q Weinberger, David Bindel, and Andrew~G Wilson.
\newblock Gpytorch: Blackbox matrix-matrix {Gaussian} process inference with {GPU} acceleration.
\newblock {\em Advances in neural information processing systems}, 31, 2018.

\bibitem{dscribe}
Lauri Himanen, Marc~OJ J{\"a}ger, Eiaki~V Morooka, Filippo~Federici Canova, Yashasvi~S Ranawat, David~Z Gao, Patrick Rinke, and Adam~S Foster.
\newblock {DScribe}: Library of descriptors for machine learning in materials science.
\newblock {\em Computer Physics Communications}, 247:106949, 2020.

\bibitem{heid2023characterizing}
Esther Heid, Charles~J McGill, Florence~H Vermeire, and William~H Green.
\newblock Characterizing uncertainty in machine learning for chemistry.
\newblock {\em Journal of Chemical Information and Modeling}, 63(13):4012--4029, 2023.

\bibitem{bishop}
Christopher~M Bishop and Nasser~M Nasrabadi.
\newblock {\em Pattern recognition and machine learning}, volume~4.
\newblock Springer, 2006.
\newblock Pages 147--152.

\bibitem{uncertainty_toolbox}
Youngseog Chung, Ian Char, Han Guo, Jeff Schneider, and Willie Neiswanger.
\newblock Uncertainty toolbox: an open-source library for assessing, visualizing, and improving uncertainty quantification.
\newblock {\em arXiv preprint arXiv:2109.10254}, 2021.

\bibitem{md17}
Stefan Chmiela, Huziel~E Sauceda, Igor Poltavsky, Klaus-Robert M{\"u}ller, and Alexandre Tkatchenko.
\newblock {sGDML}: Constructing accurate and data efficient molecular force fields using machine learning.
\newblock {\em Computer Physics Communications}, 240:38--45, 2019.

\bibitem{rmd17}
Anders~S Christensen and O~Anatole Von~Lilienfeld.
\newblock On the role of gradients for machine learning of molecular energies and forces.
\newblock {\em Machine Learning: Science and Technology}, 1(4):045018, 2020.

\bibitem{ws22}
Max Pinheiro~Jr, Shuang Zhang, Pavlo~O Dral, and Mario Barbatti.
\newblock {WS22} database, {Wigner} sampling and geometry interpolation for configurationally diverse molecular datasets.
\newblock {\em Scientific Data}, 10(1):95, 2023.

\bibitem{chemspider}
Harry~E. Pence and Antony Williams.
\newblock {ChemSpider}: An online chemical information resource.
\newblock {\em Journal of Chemical Education}, 87(11):1123--1124, 2010.

\bibitem{dftb+}
Balint Aradi, Ben Hourahine, and Th~Frauenheim.
\newblock {DFTB+}, a sparse matrix-based implementation of the {DFTB} method.
\newblock {\em The Journal of Physical Chemistry A}, 111(26):5678--5684, 2007.

\bibitem{hullermeier2021aleatoric}
Eyke H{\"u}llermeier and Willem Waegeman.
\newblock Aleatoric and epistemic uncertainty in machine learning: An introduction to concepts and methods.
\newblock {\em Machine learning}, 110(3):457--506, 2021.

\end{thebibliography}

\newpage
\renewcommand{\thepage}{S\arabic{page}} 
\renewcommand{\thesection}{S\arabic{section}}  
\renewcommand{\thetable}{S\arabic{table}}  
\renewcommand{\thefigure}{S\arabic{figure}} 
\setcounter{section}{0}
\setcounter{figure}{0}
\setcounter{page}{1}

\section{Hyperparameter optimization}\label{si:hyperparameter_optimization}
To optimize the GPR hyperparameters we maximize the marginal log likelihood (MLL) using GPyTorch\cite{gpytorch}. We use the ADAM optimizer with a learning rate of $lr=0.1$ and 200 steps. Since the initial guess of the hyperparameters in the optimization process can significantly influence the outcome, we perform cross validation for all combinations of possible initial guesses shown in Table \ref{tab:hyper_inits}. In all cases, we use 1000 random samples to optimize the hyperparameters, for which we perform 5-fold cross validation with 5 repetitions. In every fold, we optimize the hyperparameters with the respective training samples and starting values. Then we, choose the hyperparameter initial guesses, which resulted in the lowest loss on average over all repetitions and folds, and use them as intitial guesses for maximizing the MLL with all 1000 random samples.

\begin{table}[htb!]
    \centering
    \renewcommand{\arraystretch}{1.2}
    \begin{tabular}{c|c}
        Hyperparameter & Possible values\\ \hline
         Lengthscales $l_1,...,l_D$& 2, $10^{0.75}$, $10^{1.5}$\\
         Outputscale $\sigma_f$& 1\\
         Noise $\sigma^2_n$& $10^{-4}$, $10^{-6}$, $10^{-8}$ \\
    \end{tabular}
    \caption{Possible initial guesses for hyperparameters for the maximization of the MLL.}
    \label{tab:hyper_inits}
\end{table}

\newpage
\section{Ensemble mean vs single model mean}\label{si:ensemble_vs_single_mean}

\begin{figure}[htb!]
    \centering
    \begin{subfigure}[b]{\textwidth}
        \includegraphics[width=\textwidth]{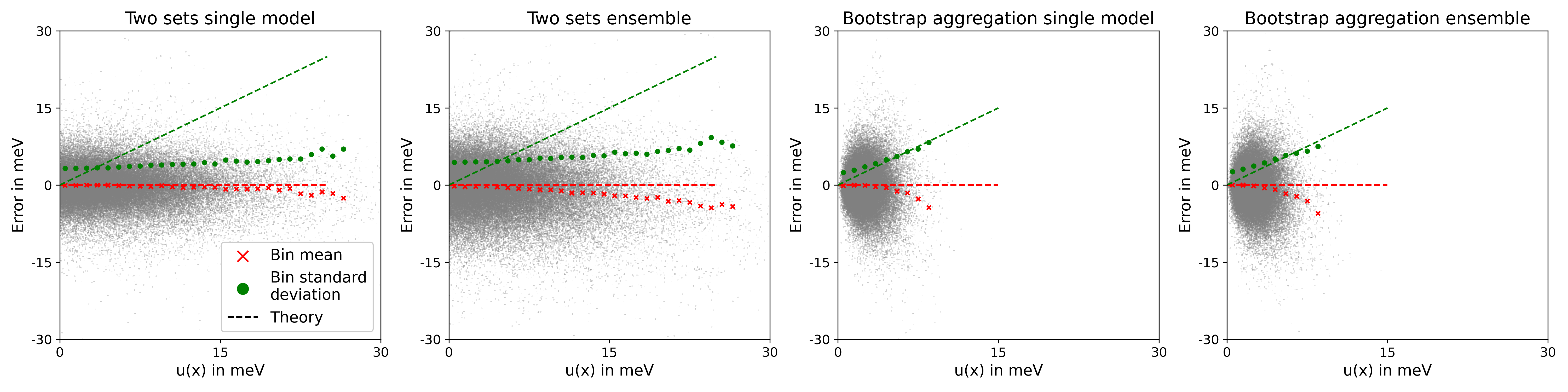}
        \caption{GPR with Coulomb}
    \end{subfigure}
    \\
    \begin{subfigure}[b]{\textwidth}
        \includegraphics[width=\textwidth]{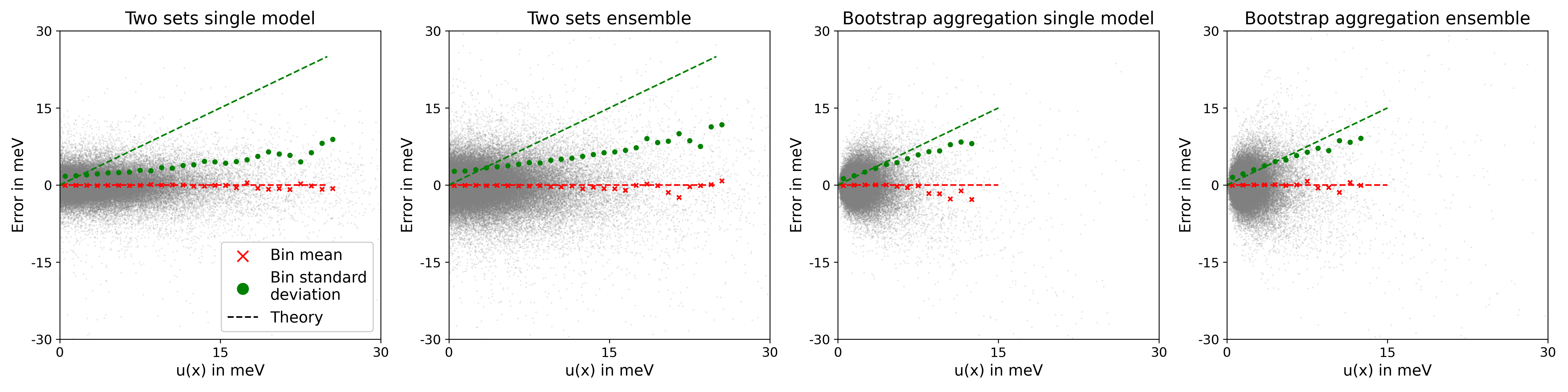}
        \caption{GPR with SOAP}
    \end{subfigure}
    \caption{Extended reliability diagrams of two sets and bootstrap aggregation uncertainty for GPR with Coulomb and SOAP representations for benzene from the rMD17 dataset.}
    \label{fig:ensemble_vs_single_rmd17_benzene}
\end{figure}

\begin{figure}[htb!]
    \centering
    \begin{subfigure}[b]{\textwidth}
        \includegraphics[width=\textwidth]{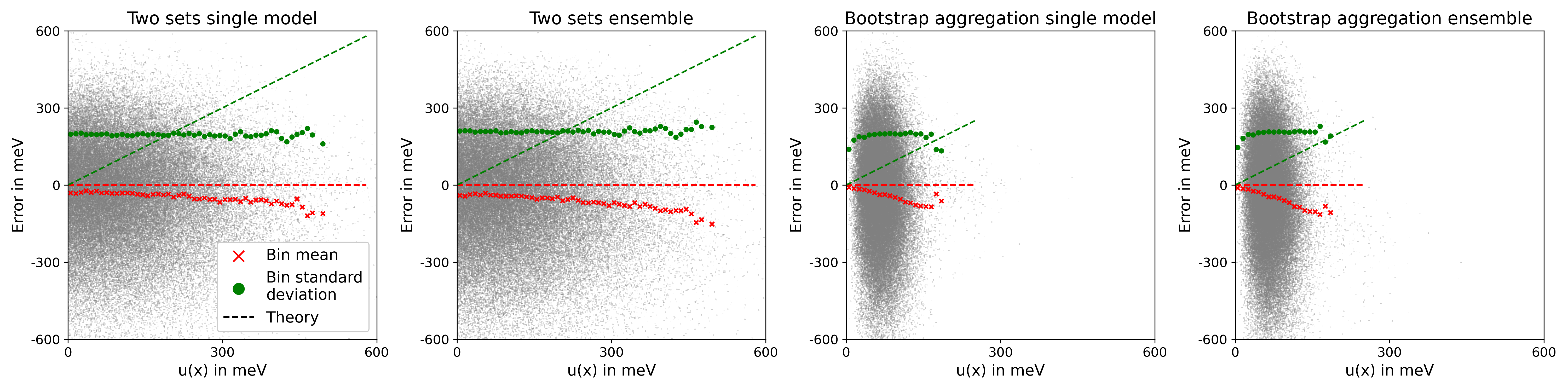}
        \caption{GPR with Coulomb}
    \end{subfigure}
    \\
    \begin{subfigure}[b]{\textwidth}
        \includegraphics[width=\textwidth]{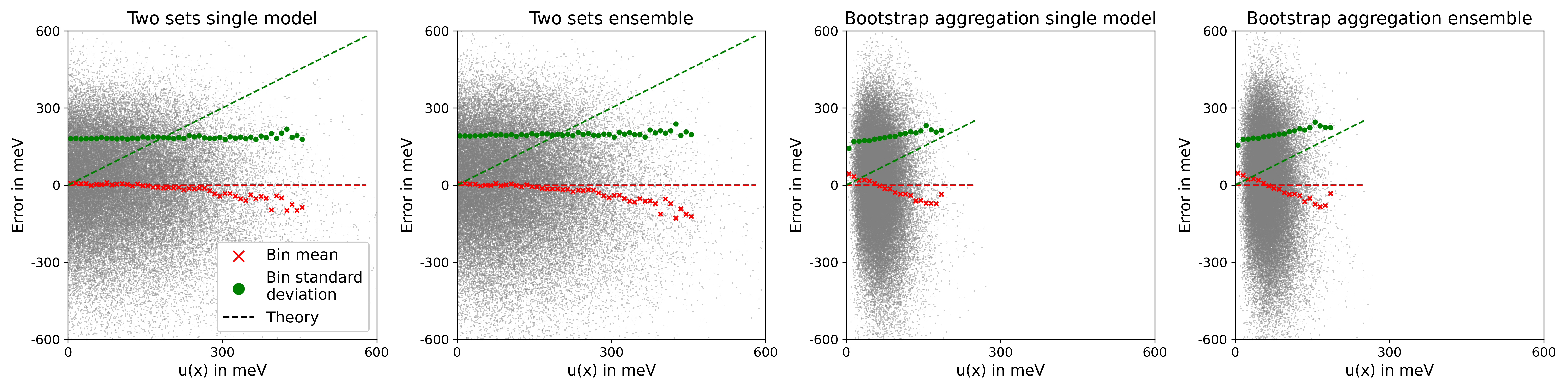}
        \caption{GPR with SOAP}
    \end{subfigure}
    \caption{Extended reliability diagrams of two sets and bootstrap aggregation uncertainty for GPR with Coulomb and SOAP representations for aspirin from the rMD17 dataset.}
    \label{fig:ensemble_vs_single_rmd17_aspirin}
\end{figure}

\newpage
\section{Bin widths}\label{si:bin_width}
\begin{table}[htb!]
    \centering
    \renewcommand{\arraystretch}{1.2}
    \begin{tabular}{c|c}
        Molecule & Bin width in meV\\ \hline
         rMD17 Benzene & 1\\
         rMD17 Aspirin & 15\\
         WS22 SMA & 20\\
         WS22 O-HBDI & 100\\
         Porphyrin & 2\\
    \end{tabular}
    \caption{Bin width of correlation plots for different molecules.}
\end{table}

\newpage

\section{Calibration analysis}
\begin{figure}[htb!]
    \centering
    \includegraphics[width=\textwidth]{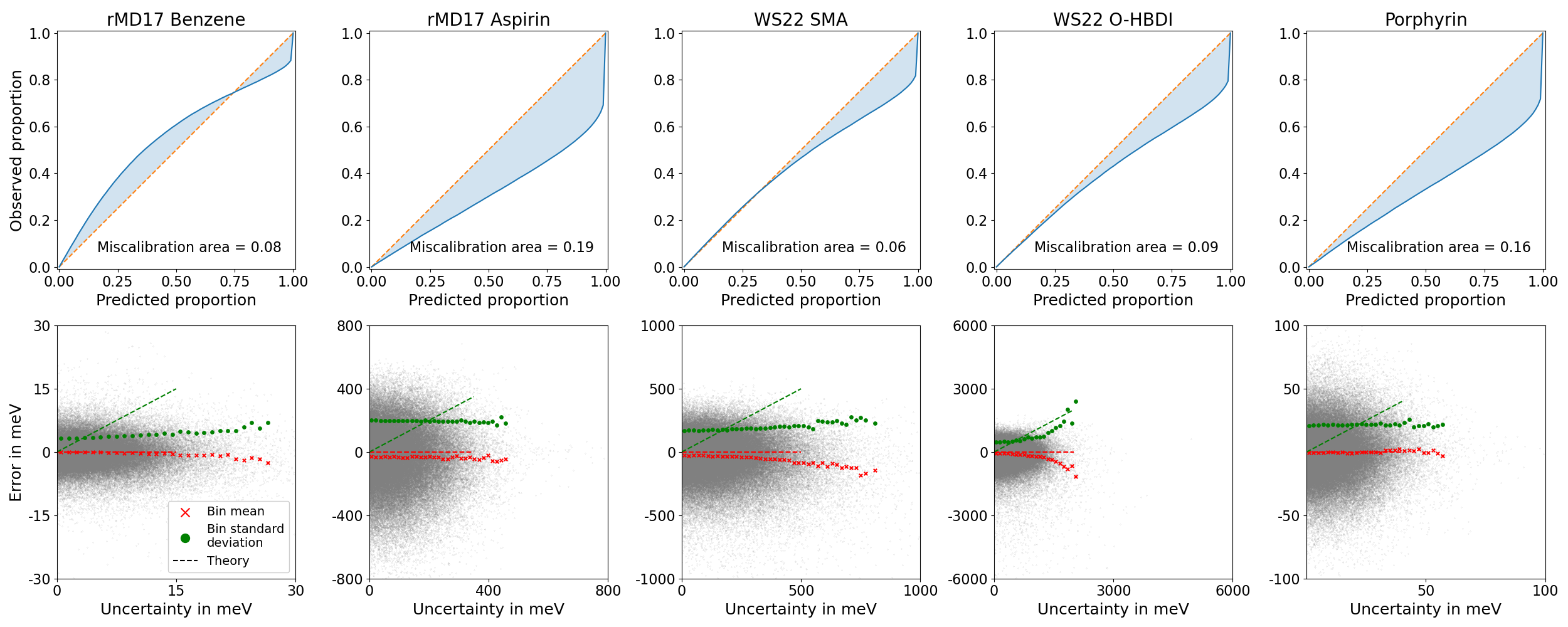}
    \caption{Calibration curves and extended reliability diagrams of the \textbf{two sets} uncertainty of GPR with \textbf{Coulomb} applied to different datasets.}
    \label{fig:calibration_coulomb_Two_sets}
\end{figure}

\begin{figure}[htb!]
    \centering
    \includegraphics[width=\textwidth]{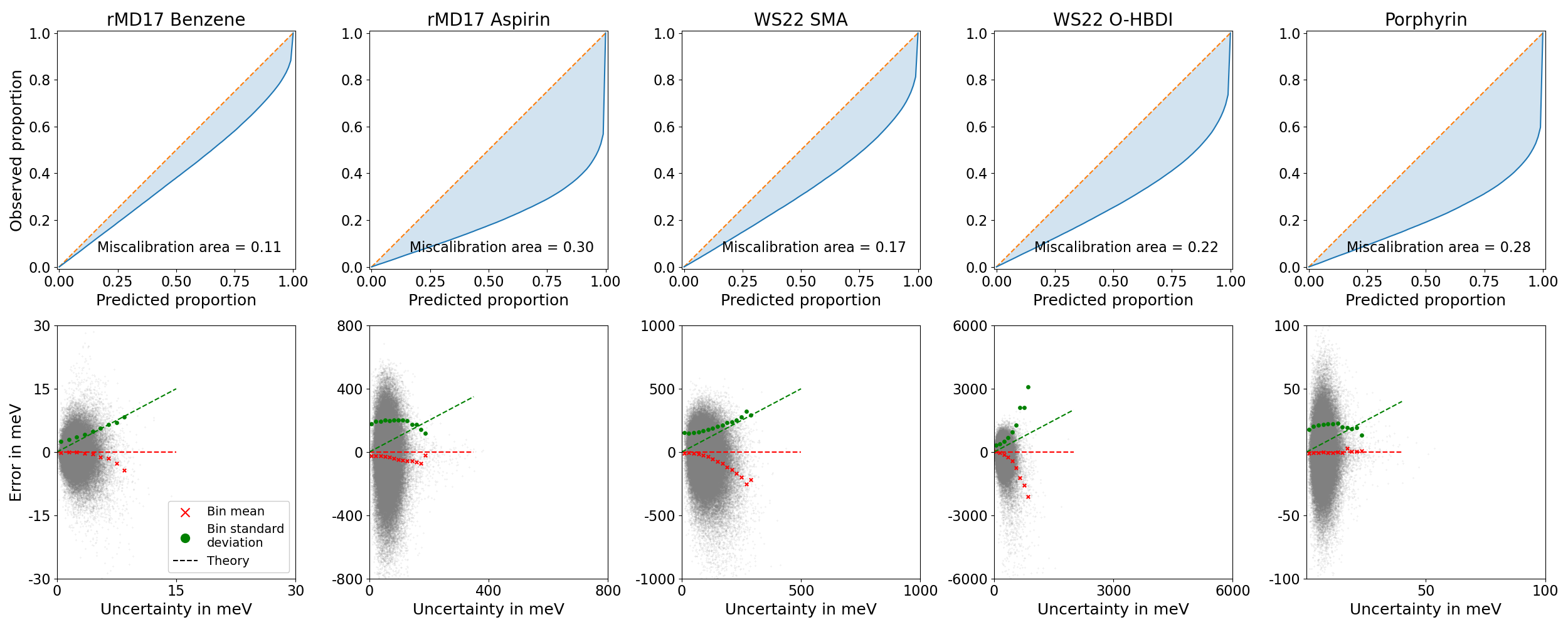}
    \caption{Calibration curves and extended reliability diagrams of the \textbf{bootstrap} uncertainty of GPR with \textbf{Coulomb} applied to different datasets.}
    \label{fig:calibration_coulomb_Bootstrap_aggregation}
\end{figure}

\begin{figure}[htb!]
    \centering
    \includegraphics[width=\textwidth]{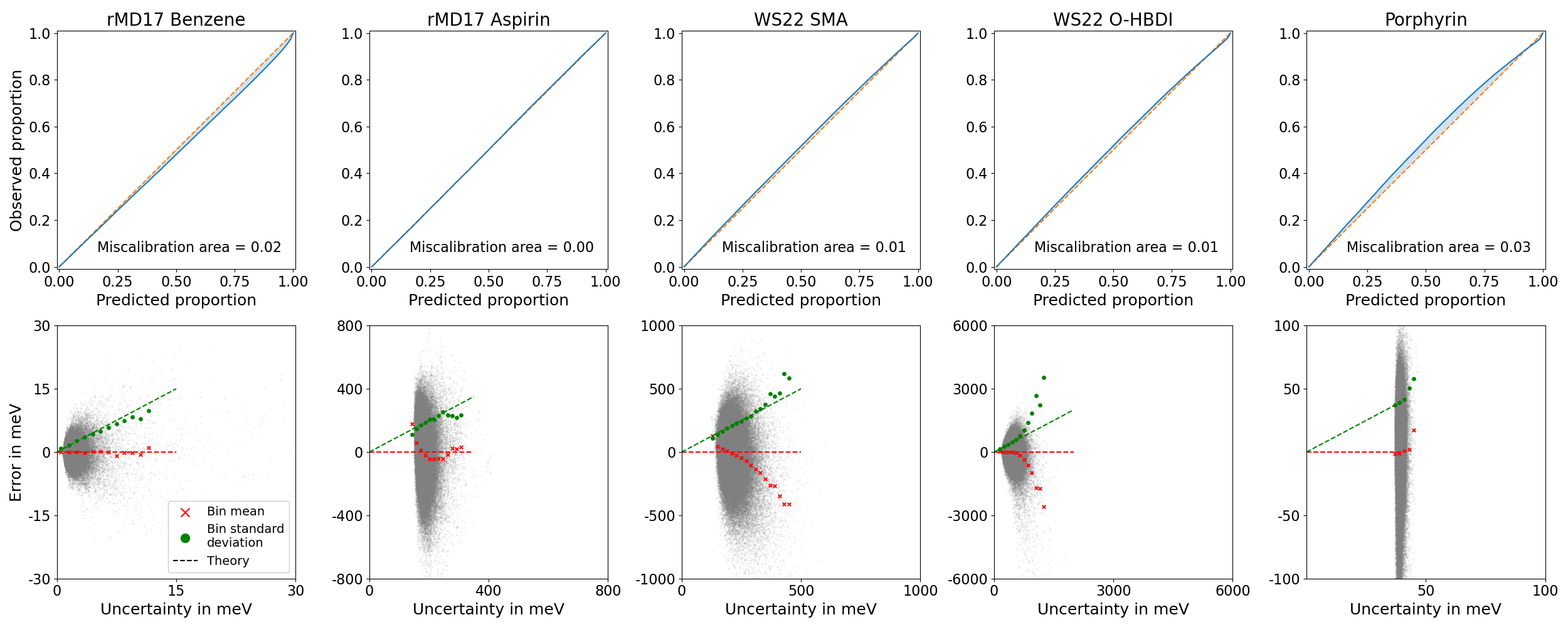}
    \caption{Calibration curves and extended reliability diagrams of the \textbf{GPR standard deviation} of GPR with \textbf{SOAP} applied to different datasets. }
    \label{fig:calibration_soap_GPR_model}
\end{figure}

\begin{figure}[htb!]
    \centering
    \includegraphics[width=\textwidth]{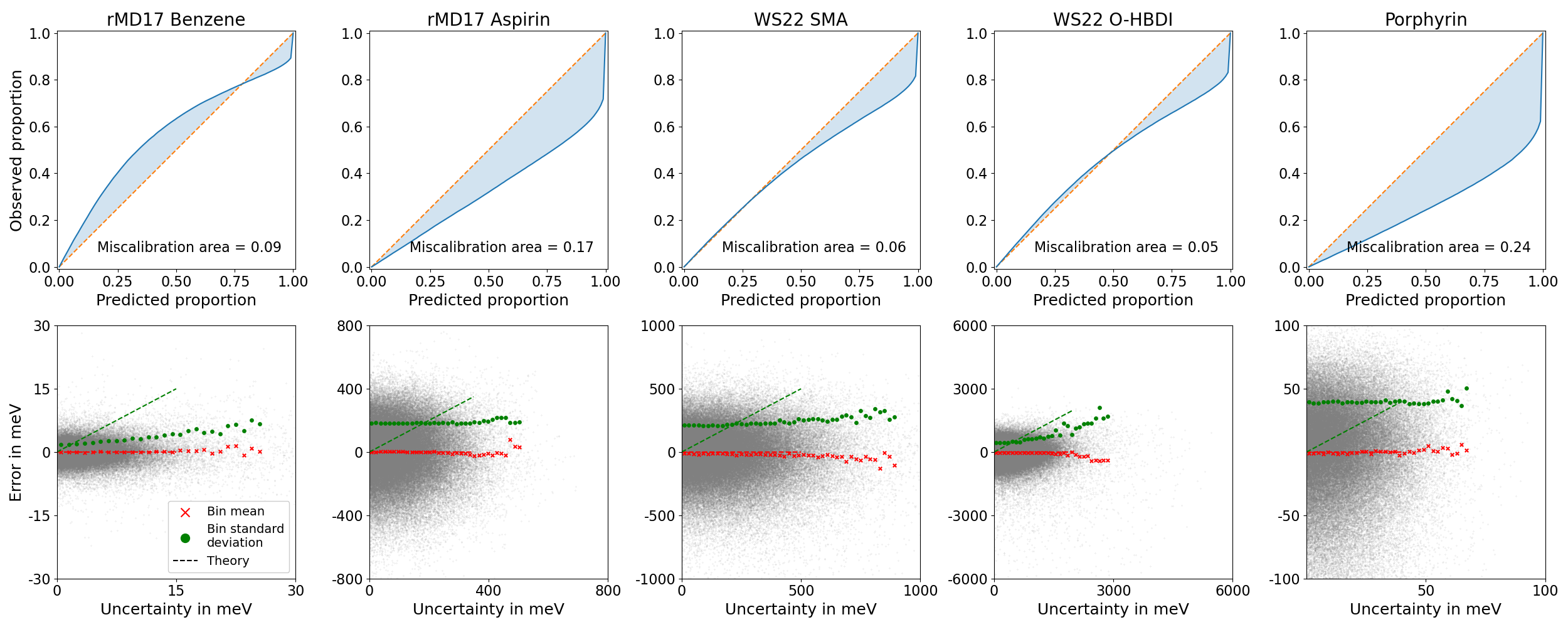}
    \caption{Calibration curves and extended reliability diagrams of the \textbf{two sets} uncertainty of GPR with \textbf{SOAP} applied to different datasets.}
    \label{fig:calibration_soap_Two_sets}
\end{figure}

\begin{figure}[htb!]
    \centering
    \includegraphics[width=\textwidth]{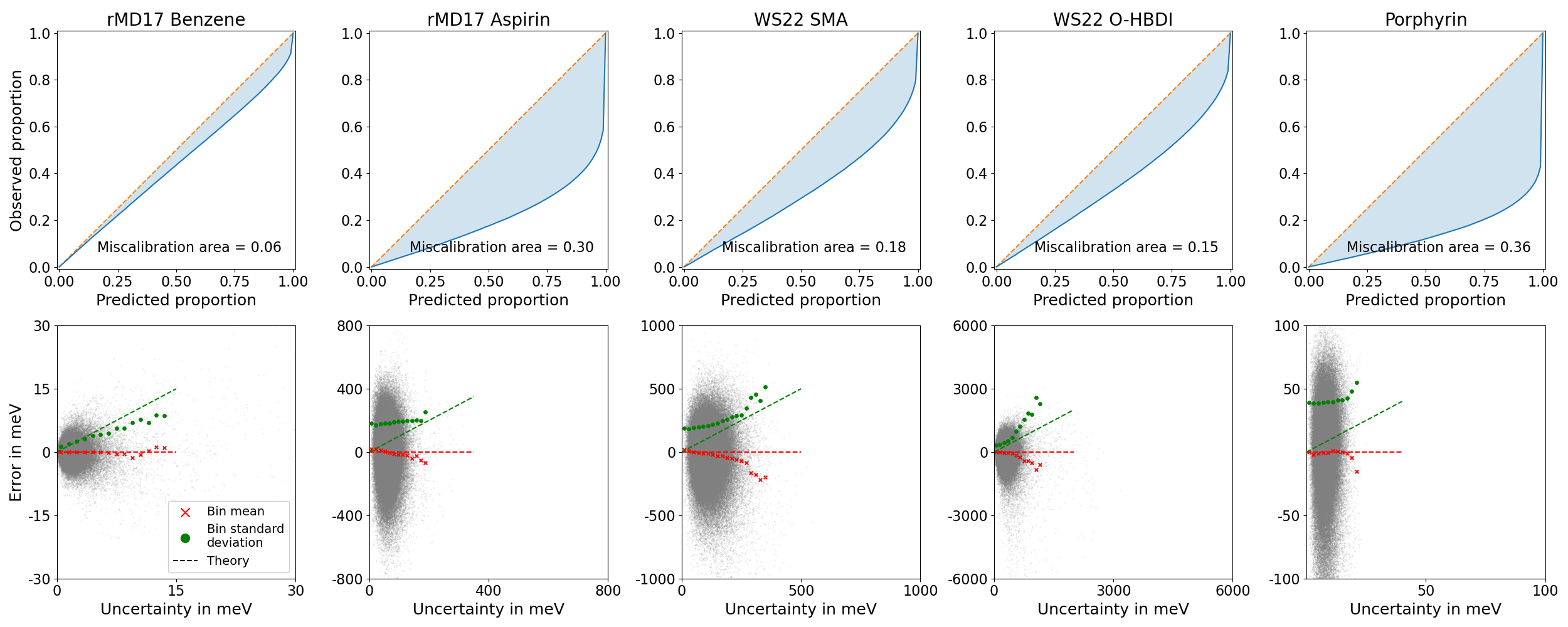}
    \caption{Calibration curves and extended reliability diagrams of the \textbf{bootstrap} uncertainty of GPR with \textbf{SOAP} applied to different datasets.}
    \label{fig:calibration_soap_Bootstrap_aggregation}
\end{figure}

\clearpage

\section{Explicit error distributions of separate bins}
\begin{figure}[htb!]
    \centering
    \begin{subfigure}[b]{\textwidth}
        \centering
        \includegraphics[width=0.8\textwidth]{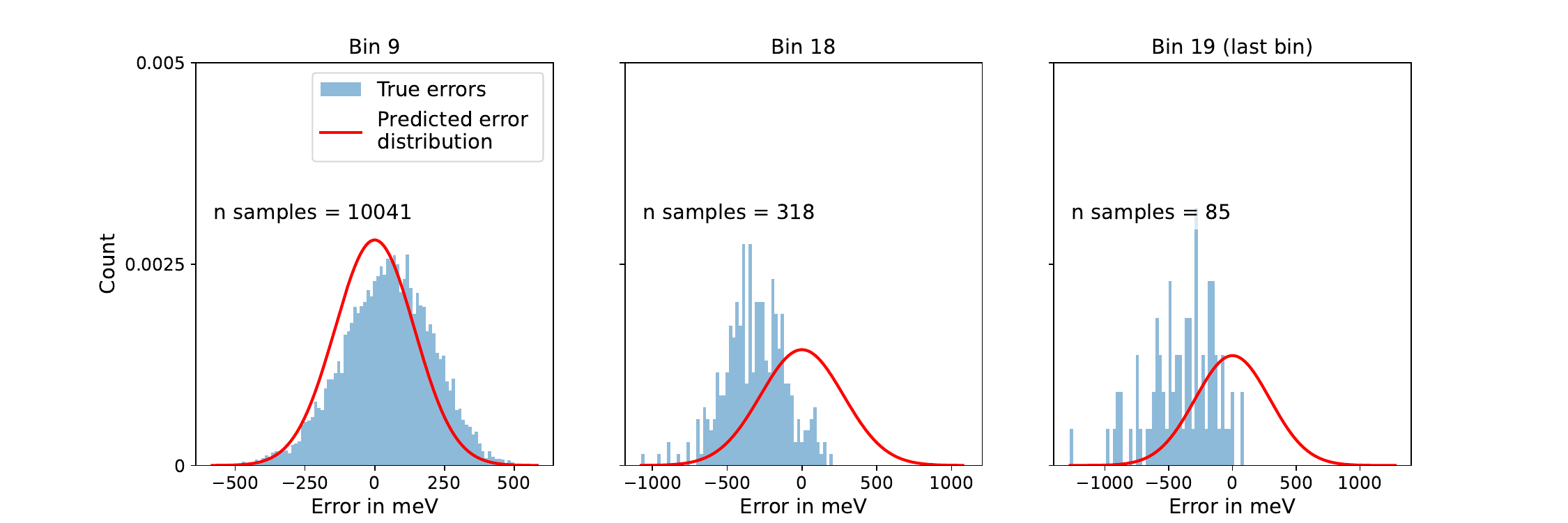}
        \caption{Aspirin}
    \end{subfigure}
    \\
    \begin{subfigure}[b]{\textwidth}
        \centering
        \includegraphics[width=0.8\textwidth]{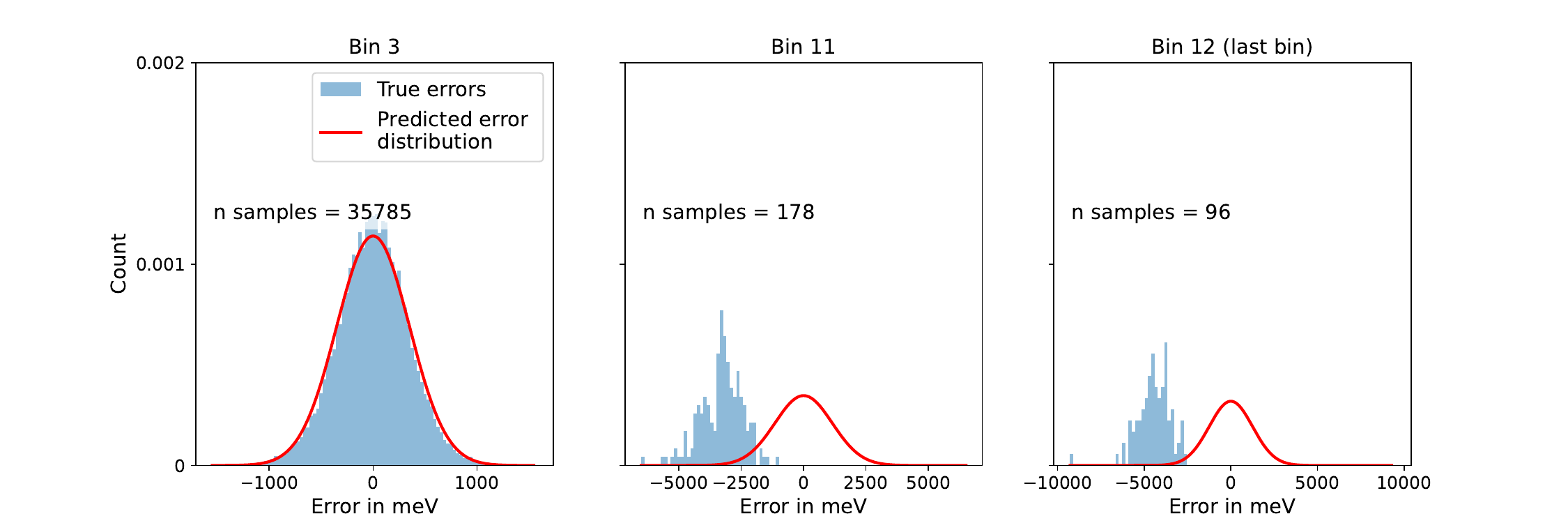}
        \caption{O-HBDI}
    \end{subfigure}
    \caption{Predicted error distribution and distribution of true errors for selected bins of the reliability diagram of GPR with SOAP for Aspirin and O-HBDI.}
    \label{fig:single_bin_dist}
\end{figure}

\newpage
\newpage
\section{Other metrics uncertainty sampling}
\begin{figure}[htb!]
    \centering
    \begin{subfigure}[b]{\textwidth}
        \includegraphics[width=\textwidth]{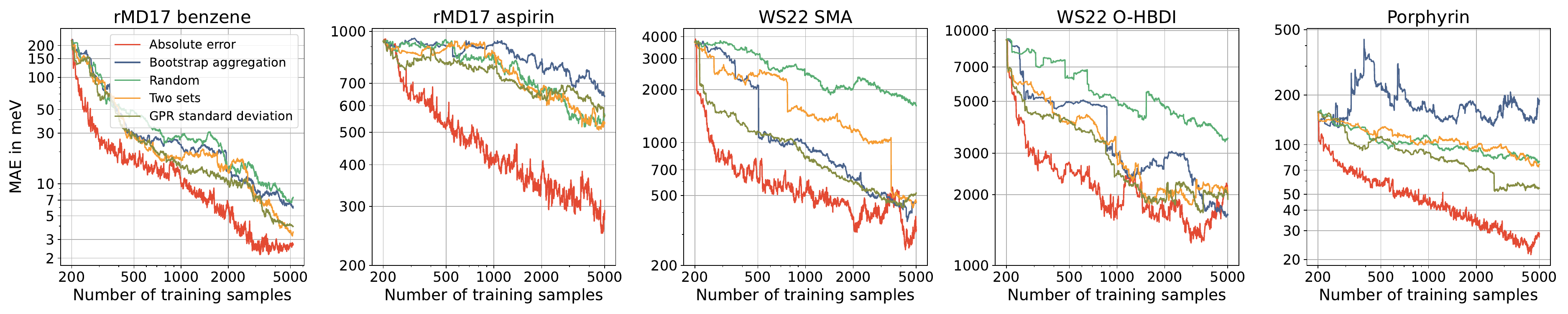}
        \caption{GPR with Coulomb}
    \end{subfigure}
    \\
    \begin{subfigure}[b]{\textwidth}
        \centering
        \includegraphics[width=0.8\textwidth]{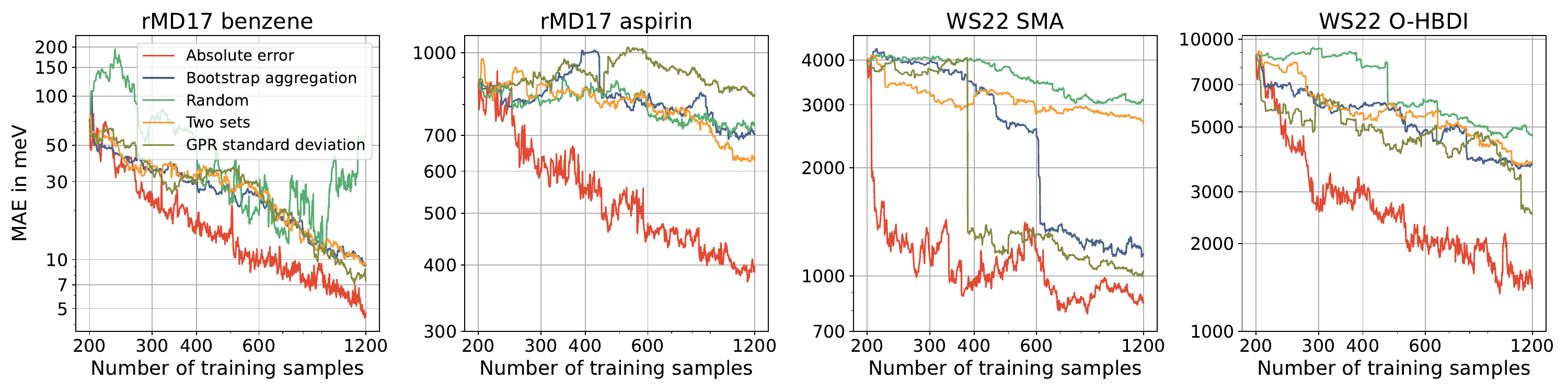}
        \caption{GPR with SOAP}
    \end{subfigure}
    \caption{Maximum absolut error of uncertainty sampling for GPR with Coulomb and SOAP representations for different datasets. }
    \label{fig:uncertainty_sampling_max}
\end{figure}

\begin{figure}[htb!]
    \centering
    \begin{subfigure}[b]{\textwidth}
        \includegraphics[width=\textwidth]{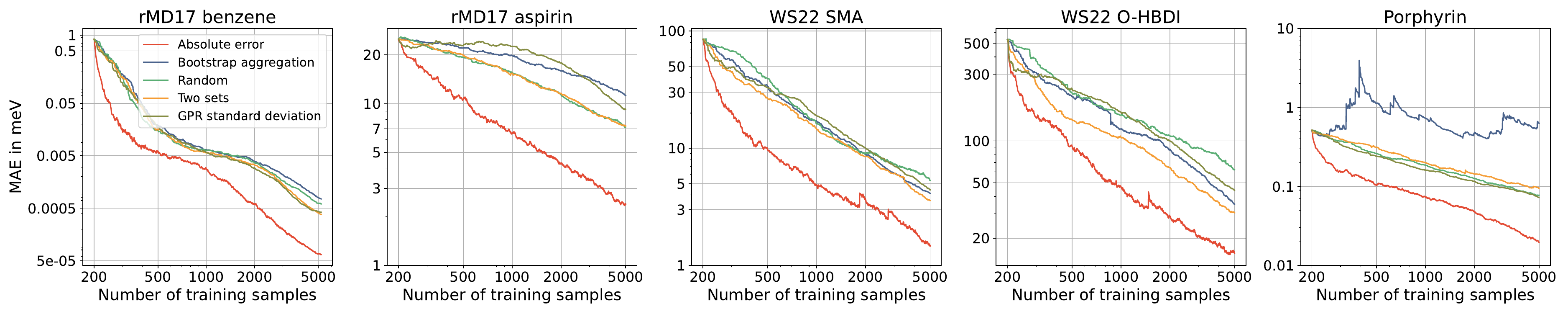}
        \caption{GPR with Coulomb}
    \end{subfigure}
    \\
    \begin{subfigure}[b]{\textwidth}
        \centering
        \includegraphics[width=0.8\textwidth]{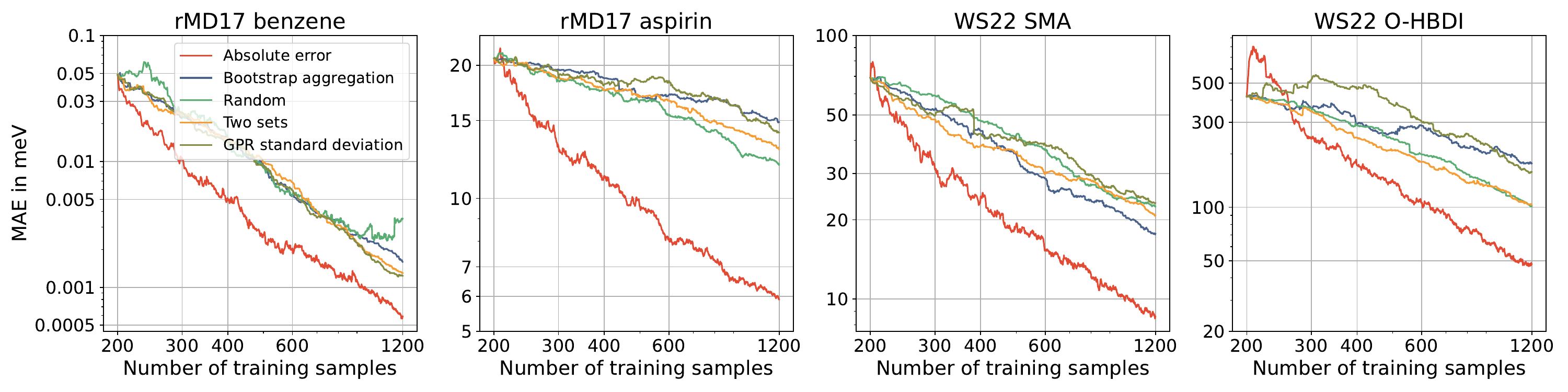}
        \caption{GPR with SOAP}
    \end{subfigure}
    \caption{Test variance of uncertainty sampling for GPR with Coulomb and SOAP representations for different datasets.}
    \label{fig:uncertainty_sampling_var}
\end{figure}
\end{document}